\documentclass{article}

\usepackage[preprint]{neurips_2023}

\usepackage[utf8]{inputenc} 
\usepackage[T1]{fontenc}    
\usepackage[colorlinks,citecolor=blue,linkcolor=magenta,bookmarks=true,hypertexnames=false,backref=page]{hyperref}       
\usepackage{url}            
\usepackage{booktabs}       
\usepackage{caption}  
\usepackage{array}    
\usepackage{amsfonts}       
\usepackage{nicefrac}       
\usepackage{microtype}      
\usepackage{xcolor}         
\usepackage{xspace}

\usepackage{graphicx}
\usepackage{subcaption}
\usepackage{lipsum}
\usepackage{pifont}
\usepackage[most]{tcolorbox}
\newtcolorbox{highlight}[1][]{
    enhanced,
    colback=yellow!10,
    colframe=gray!30,
    boxrule=0.5pt,
    arc=2pt,
    leftrule=3pt,
    rightrule=3pt,
    toprule=1pt,
    bottomrule=1pt,
    breakable,
    #1
}

\usepackage{amsmath}
\usepackage{amssymb}
\usepackage{mathtools}
\usepackage{amsthm}
\usepackage{float}

\newcommand{\bvec}[1]{\mathbf{#1}}
\newcommand{\vanilla}{{\fontfamily{cmss}\selectfont Standard Hybrid}\xspace}
\newcommand{\variant}{{\fontfamily{cmss}\selectfont MambaFormer}\xspace}

\makeatletter
\def\blfootnote{\xdef\@thefnmark{}\@footnotetext}
\makeatother

\newcommand{\new}[1]{{\color{black} #1}}

\usepackage[capitalize,noabbrev]{cleveref}

\theoremstyle{plain}

\theoremstyle{definition}

\theoremstyle{remark}

\usepackage[textsize=tiny]{todonotes}

\title{Can Mamba Learn How to Learn?\\ A Comparative Study on In-Context Learning Tasks}

\author{
Jongho Park$^1$,
~Jaeseung Park$^{2}$\thanks{This work was done during an internship at KRAFTON.},
~Zheyang Xiong$^3$,
~Nayoung Lee$^3$,
~Jaewoong Cho$^1$,\\
~\textbf{Samet Oymak$^4$,}
~\textbf{Kangwook Lee$^{1,3}$,}
~\textbf{Dimitris Papailiopoulos$^{1,3}$}\\
\\
$^1$ KRAFTON, $^2$ Seoul National University,\\
$^3$ University of Wisconsin-Madison, 
$^4$ University of Michigan, Ann Arbor
}

\hypersetup{
    pdftitle={Can Mamba Learn How to Learn? A Comparative Study on In-Context Learning Tasks},
}

\begin{document}

\maketitle

\begin{abstract}
    State-space models (SSMs), such as Mamba~\citep{Gu2023mamba}, have been proposed as alternatives to Transformer networks in language modeling, by incorporating gating, convolutions, and input-dependent token selection to mitigate the quadratic cost of multi-head attention. Although SSMs exhibit competitive performance, their in-context learning (ICL) capabilities, a remarkable emergent property of modern language models that enables task execution without parameter optimization, remain underexplored compared to Transformers. In this study, we evaluate the ICL performance of SSMs, focusing on Mamba, against Transformer models across various tasks. Our results show that SSMs perform comparably to Transformers in standard regression ICL tasks, while outperforming them in tasks like sparse parity learning. However, SSMs fall short in tasks involving non-standard retrieval functionality. To address these limitations, we introduce a hybrid model, \variant, that combines Mamba with attention blocks, surpassing individual models in tasks where they struggle independently. Our findings suggest that hybrid architectures offer promising avenues for enhancing ICL in language models.
\end{abstract}

\blfootnote{Email: \texttt{<jongho.park@krafton.com>}. Correspondence: \texttt{<dimitris@papail.io>}}

\section{Introduction}\label{sec:intro}

Modern large language models (LLMs) exhibit remarkable in-context learning (ICL) capabilities, enabling them to learn new tasks with a few demonstrations and without further weight fine-tuning. Although the exact emergence mechanism of these capabilities warrants further theoretical and empirical investigation~\citep{Chan2022Data, wei2022emergent, Min2022Rethinking, Schaeffer2023Mirage}, experiments on larger Transformer-based models consistently demonstrate that their ICL capabilities improve as training loss reduces~\citep{brown2020language, Kaplan2020Scaling, Muennighoff2023Scaling}.

Meta-learning, or ``learning to learn,'' has been extensively studied~\citep{Schmidhuber1997Shifting, Ravi2016Optimization} and recently regained interest in the context of ICL, particularly concerning Transformer models~\citep{Vaswani2017Attention}. \citet{Garg2022Transformers}, for example, proposed various ICL tasks, such as learning linear regression, and evaluated the ability of transformers to perform them when specifically trained to do so. On the other hand, \citet{Min2022Metaicl} studied fine-tuning language models to explicitly learn and perform ICL. Following these footsteps, numerous research studies have been dedicated to understanding the mechanics of Attention that enable such meta-learning capabilities, either through constructive arguments or extensive experimental investigation~\citep{Akyurek2022InContextLearning, Liu2022Transformers, Bai2023Statistician, Giannou2023Looped, li2023transformers, Oswald2023aTransformersGradient, Oswald2023bMesaOptimization, yang2023looped, Zhou2023Algorithms}.

As Transformer language models are currently the only large models that have been reported to be capable of ICL in practice, this raises the question:

\begin{center} 
\fbox{ 
  \begin{minipage}{0.4\textwidth} 
  \centerline
    {\it Can attention-free models perform ICL?}
  \end{minipage}
}
\end{center}

\definecolor{green}{RGB}{1, 152, 117}
\begin{table}[t!]
\fontsize{10pt}{10pt}\selectfont
\centering
\begin{tabular}{@{}lccc@{}}
\toprule
                                & Transformer      & Mamba            & \variant  \\ \midrule
Linear regression               & \textcolor{green}{\ding{51}}          & \textcolor{green}{\ding{51}}          & \textcolor{green}{\ding{51}}      \\
Sparse linear regression        & \textcolor{green}{\ding{51}}          & \textcolor{green}{\ding{51}}          & \textcolor{green}{\ding{51}}      \\
2NN regression                  & \textcolor{green}{\ding{51}}          & \textcolor{green}{\ding{51}}          & \textcolor{green}{\ding{51}}      \\
Decision Tree                   & \textcolor{green}{\ding{51}}          & $\blacktriangle$ & \textcolor{green}{\ding{51}}      \\
Orthogonal-outlier regression   & \textcolor{green}{\ding{51}}          & $\blacktriangle$ & \textcolor{green}{\ding{51}}      \\
Many-outlier regression         & $\blacktriangle$ & \textcolor{green}{\ding{51}}          & \textcolor{green}{\ding{51}}      \\
Sparse parity                   & \textcolor{red}{\ding{55}}          & \textcolor{green}{\ding{51}}          & \textcolor{green}{\ding{51}}      \\
Chain-of-Thought I/O            & \textcolor{green}{\ding{51}}          & \textcolor{green}{\ding{51}}          & \textcolor{green}{\ding{51}}      \\
Vector-valued MQAR                 & \textcolor{green}{\ding{51}}          & \textcolor{red}{\ding{55}}          & \textcolor{green}{\ding{51}}      \\
\hfill
\end{tabular}
\caption{Model performances on various ICL tasks. We label the model's performance with \textcolor{green}{\ding{51}} if the model performs on par with other baseline models, \textcolor{red}{\ding{55}} if the model struggles to learn the task, and  $\blacktriangle$ if the performance improves but with a performance gap compared to other baseline models. Transformer fails in learning sparse parity, showing performance no better than random guessing, while Mamba suffers to accurately retrieve the value vector in vector-valued MQAR. Our proposed \variant performs on par with other baseline models in all tasks.}
\label{tab:tasks}
\end{table}

This question holds merit, especially considering that several recent studies have attempted to move beyond attention-based networks due to their quadratic cost~\citep{Katharopoulos2020transformers, Zhai2021Attention, Dao2022HungryHippos, Poli2023HyenaHierarchy, Peng2023RWKV, Sun2023RetentiveNetwork, Yang2023GLA}. 
In this work, we focus specifically on state-space models (SSMs), and particularly Mamba~\citep{Gu2023mamba}. Mamba was recently demonstrated to be highly efficient while achieving near state-of-the-art performance in standard pretraining language data sets, such as the Pile \citep{gao2020pile}, but at smaller model scales (\emph{e.g.}, up to 3 billion parameters), surpassing transformers and other attention-free architectures across various language and non-language tasks. However, ICL capabilities usually emerge at scales beyond 3 billion parameters. As a result, the potential of these attention-free models to perform ICL remains underexplored, as testing such hypotheses usually requires scaling beyond the 7 billion parameter level. Nonetheless, we can still investigate small-scale ICL capabilities by specifically training a model to perform in-context learning, following the approach of \citet{Garg2022Transformers}. 

\paragraph{Contributions.}

In this study, we introduce a diverse set of ICL tasks to evaluate the performance of Transformer and various SSMs, including state-of-the-art models like Mamba and S4 \citep{Gu2022structured}. Our findings reveal that most of these SSMs can effectively perform ICL, matching the performance of Transformers across multiple tasks. However, Mamba demonstrates some limitations in learning decision trees and retrieval tasks, but can outperform Transformers in other complex ICL tasks, such as sparse parity, where Transformer models struggle. Performance of different models on each task is summarized in \cref{tab:tasks}.

\begin{figure}[t!]
  \centering\includegraphics[width=0.7\textwidth]{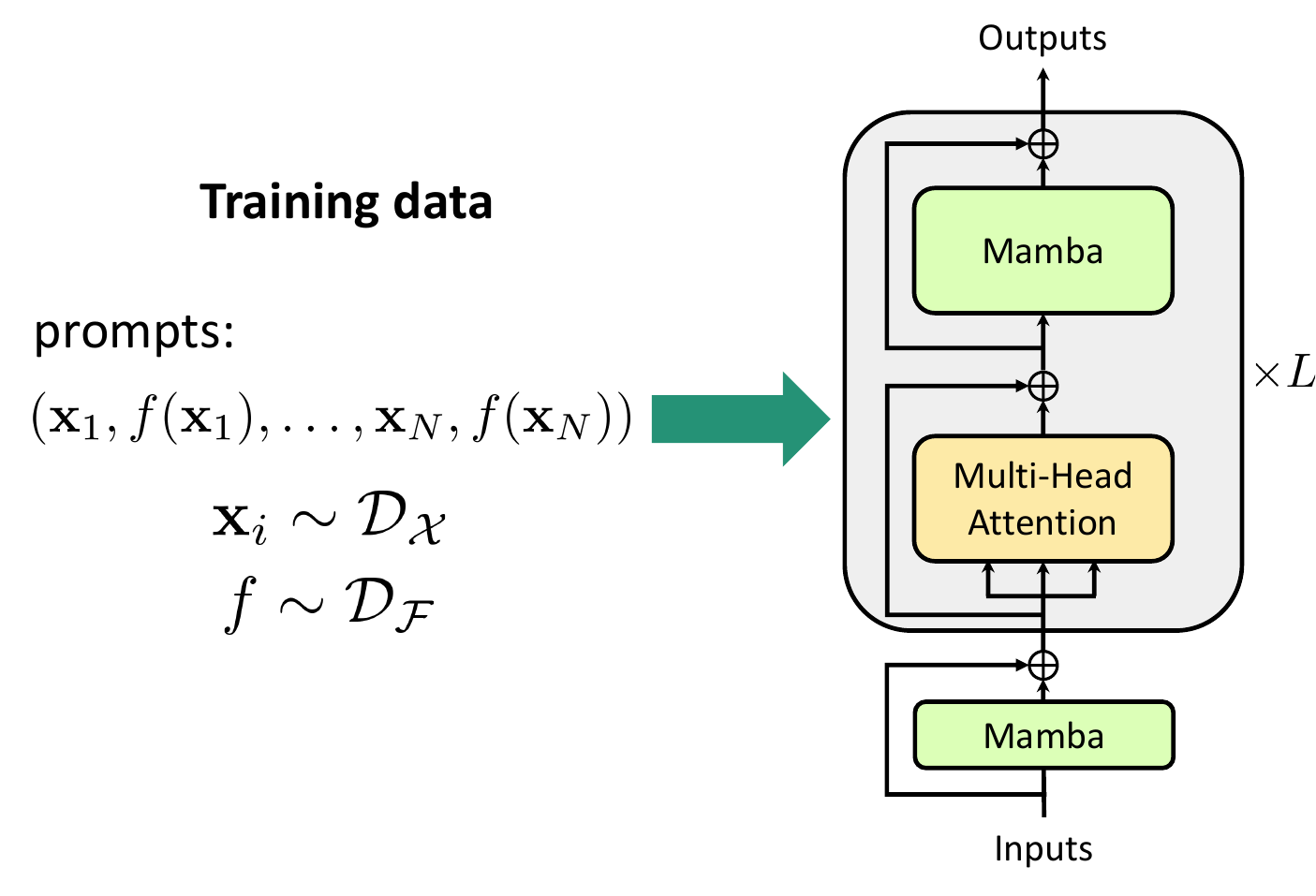}
  \caption{\variant is a hybrid architecture that replaces MLP blocks within the transformer with Mamba blocks. Importantly, the architecture starts with a Mamba block and does not use positional encoding. In our ICL evaluations, we find that \variant consistently achieves a best-of-both-worlds performance when compared to Transformer and Mamba.}
  \label{fig:mambaF}
\end{figure}

Since there seem to be tasks where either family of models is better, we explore the impact of interleaving  SSM blocks with multi-head attention blocks, similar to \citep{Gu2023mamba}. We introduce \variant, a novel hybrid architecture that integrates Mamba and Attention layers, while eliminating the need for positional encodings, as shown in \cref{fig:mambaF}. \variant seems to leverage the strengths of both Mamba and Transformers, exhibiting good performance across all evaluated  ICL tasks and simultaneously learning sparse parity and retrieval.
\footnote{Code is available at \url{https://github.com/krafton-ai/mambaformer-icl}.}

We believe that our findings underscore the importance of broadening the understanding of ICL beyond Transformers, as significant progress has been made in the context of attention-free architectures.

We acknowledge that a limitation of our study lies in the focus on non-language ICL tasks and smaller models. It is possible that an architectural comparison between SSMs and transformers for more general ICL tasks in actual language settings at higher parameter counts might not be yield the same observations as we offer here. 
Nevertheless, \new{we show potential ICL language capabilities of these architectures by conducting experiments on synthetic formal language ICL datasets~\citep{Xie2021Explanation, Akyurek2024ICLL}.}
Moreover, our non-language empirical findings indicate that, apart from its difficulty in some retrieval tasks, similar to those noted by \citet{Arora2023Zoology}, there seems to be no fundamental obstacle for Mamba to perform in-context learning.

\section{Related Work}\label{sec:related_work}


\paragraph{Transformer-based in-context learning.}
The role of attention in ICL has been the focus of both theoretical and empirical research.
Studies have primarily focused on  meta-learning~\citep{Ravi2016Optimization, Min2022Metaicl}, where one explicitly trains for ICL.
Notably, \citet{Garg2022Transformers} have examined transformers in in-context regression tasks, from learning linear regression to learning decision trees.
Subsequent works have suggested that attention may mimic various optimization algorithms~\citep{Akyurek2022InContextLearning, Oswald2023bMesaOptimization, Dai2023GPTMetaOptimizers}.
In fact,~\citet{Ahn2023Transformers, Mahankali2023OneGD} have provably shown that 
the global minimum of the linear regression ICL objective implements one step
of preconditioned gradient descent for one layer of linear attention.

While these settings might appear simplistic and detached from language models, ~\citet{Bhattamishra2023TransformersLLMs}  showed that a frozen GPT-2 can implement the nearest neighbor algorithm, drawing connections between the ICL in existing language models and the stylized setting of training for ICL from random initialization.
Furthermore, ~\citet{Olsson2022incontext} also empirically demonstrate that ``induction heads'', which are attention heads that solve a simple retrieval problem, correlate with ICL behavior, providing a strong connection between retrieval and ICL.

\paragraph{Sub-quadratic architectures.}
The number of effective floating point operations in an attention layer scales quadratically with respect to the input sequence length. Numerous approximations or alternative model architectures have been proposed to overcome the quadratic dependence. These
range from approximating attention mechanisms~\citep{Beltagy2020Longformer, Wang2020Linformer} to the development of novel recurrent convolutional models such as structured state-space models~\citep{Gu2022structured}.

S4~\citep{Gu2022diagonal} is a family of sequence models characterized by a discretized state-space model
\begin{equation}
    \label{eq:ssm}
    \bvec h_t = \overline{\bvec A} \bvec h_{t-1} + \overline{\bvec B} \bvec x_{t}, ~y_t = \bvec C\bvec h_{t},
\end{equation}

where $\bvec h_t$ represents the hidden state and $(\overline{\bvec A}, \overline{\bvec B}, \bvec C)$ are input-independent  (transformed) parameters.
The recurrence is expressible as a convolution, enabling near-linear complexity using Fast Fourier Transform.
Viewed in this framework, Linear Transformers~\citep{Katharopoulos2020transformers}, which employ linear attention without softmax, can be seen as a variant of linear SSM.

Building upon this concept, H3~\citep{Dao2022HungryHippos} integrates an S4 with dual gated connections. 
The recent Mamba~\citep{Gu2023mamba} departs from the standard SSM by introducing a selection mechanism that makes $(\overline{\bvec A}, \overline{\bvec B}, \bvec C)$ in \cref{eq:ssm} dependent on the input $\bvec x_t$ allowing input-dependent sequence mixing.

There are other notable attention-free models such as Hyena~\citep{Poli2023HyenaHierarchy}, RWKV~\citep{Peng2023RWKV}, RetNet~\citep{Sun2023RetentiveNetwork}, and GLA~\citep{Yang2023GLA}.
Despite of state-of-the-art performance for models like Mamba, \citet{Arora2023Zoology} have demonstrated that subquadratic models still lag behind attention on multi-query recall tasks, which is a generalization of the induction head task \citep{Olsson2022incontext}.

In their study,  \citet{Xie2021Explanation} introduced a synthetic language-based dataset for in-context learning, named GINC, and demonstrated that both transformers and LSTMs~\citep{Hochreiter1997lstm} can perform ICL. Notably, LSTMs outperformed transformers in ICL accuracy on GINC, a finding similar to that found in \citet{Liu2023flipflop} for their flip-flop language modeling task.
More recently, \cite{Akyurek2024ICLL} proposed a language-based ICL benchmark for training models on formal languages generated by random finite automata. Their results showed that Transformers notably better than subquadratic models, establishing a benchmark that effectively measures ICL in language modeling.

\section{Experimental Setup}\label{sec:experiments}

We evaluate the ICL capabilities of SSMs and Transformers by training each model from scratch on each specific task, detailed in Section~\ref{sec:training}. Section~\ref{sec:tasks} outlines the ICL and related tasks investigated in our study. We provide a brief summary of our tasks in the following \cref{tab:summary_of_tasks}.

\begin{table*}[ht]
\centering
\scalebox{0.8}{
\begin{tabular}{@{}l|cccc@{}}
\toprule
Task & dim $(d)$ & points ($N$)  & Example/Function Sampling & Task-specific  \\ 
\midrule
Linear regression & 20 & 41 & $\bvec x, \bvec w \sim \mathcal{N}(0, {\bvec I}_d)$ & --\\ 
Sparse Linear regression & 20 & 101 & $\bvec x, \bvec w \sim \mathcal{N}(0, {\bvec I}_d)$,~$\texttt{sparsity}(\bvec w)\gets k$ & $k=3$ \\
2NN regression & 20 & 101 & $\bvec W^{(1)}_{ij}, \bvec W^{(2)}_{ij} \sim \mathcal{N}(0, 1)$ & -- \\ 
Decision Tree & 20 & 101 & $\bvec x, \mathrm{Leaf} \sim \mathcal{N}(0, 1), \mathrm{non\_leaf} \sim \{1,...,d\}$  & $\mathrm{depth}=4$\\
Orthogonal-outlier regression & 20 & 101 & $\bvec x, \bvec w \sim \mathcal{N}(0, {\bvec I}_d)$, $\bvec u, \bvec v \sim \bvec w^{\perp}$ &  $p=0.5$ \\
Many-outlier regression & 20 & 512 & $\bvec x \sim \mathcal{N}(0, I)$ w.p. $1-p$, else $(\bvec x, y) = (\mathbf{1}, 1)$ & $p=0.9$ \\ 
Sparse Parity & 10 & 140 & $\bvec x \sim \{-1,1\}^d, y = \prod_{j \in I} \bvec x[j]$ & $k=2$\\
Chain-of-Thought I/O & 10 & 101 & $\bvec x \sim \mathcal{N}(0, {\bvec I}_d)$, $ {\bvec W}_{ij} \sim \mathcal{N}(0, 2/k)$,  $\bvec v \sim \mathcal{N}(0, {\bvec I}_k)$ & $h=8$ \\
Vector MQAR & 20 & 128 & $\bvec k, \bvec v \sim \mathrm{Unif}(\mathcal{S}^{d-1})$ & $32$ k-v pairs\\
\bottomrule
\end{tabular}}
\caption{Summary of Tasks. All models are trained for 500,000 iterations (except for the vector MQAR; see \cref{appendix:mqar_setup}).}
\label{tab:summary_of_tasks}
\end{table*}

\subsection{Model Training for In-context Learning}\label{sec:training}
We train models to learn a specific function class $\mathcal{F}$ in-context. Training begins by generating random prompts: selecting a function $f\in\mathcal{F}$ from distribution $\mathcal{D}_{\mathcal{F}}$ and sampling a sequence of random inputs $\bvec x_1,\ldots, \bvec x_{N}\in\mathbb{R}^d$ i.i.d. from $\mathcal{D}_{\mathcal{X}}$. Here, $N$ and $d$ represent the number of in-context examples and the dimension of $\bvec x_i$, respectively. These inputs create the prompt $P=(\bvec x_1, f(\bvec x_1), \ldots, \bvec x_N, f(\bvec x_N))$.
We train the model ${f}_{\theta}$, parameterized by $\theta$, by minimizing the expected loss over all prompts: 
\begin{equation}
\label{eq:objective}
    \min_\theta \mathbb{E}_{P}\left[\frac{1}{N}\sum_{i=1}^{N-1}\ell(f_{\theta}(P^i),f(\bvec x_i))\right], 
\end{equation}
where $P^i:=(\bvec x_1, f(\bvec x_1), \ldots, \bvec x_i, f(\bvec x_i), \bvec x_{i+1})$ and $\ell(\cdot, \cdot)$ is a loss function. 
Since $f: \mathbb{R}^d \rightarrow \mathbb{R}$, we append $d-1$ zeros to $f(\bvec x)$ to match the dimensions.
We use appropriate loss functions for each task.

\paragraph{Model architecture.} We primarily focus on SSMs, including (1) Mamba~\citep{Gu2023mamba}, a state-of-the-art SSM model with selection mechanism; (2) S4~\citep{Gu2022diagonal}, a linear time-invariant predecessor of Mamba; and (3) S4-Mamba, a variant where Mamba's input-dependent S6 is replaced with input-independent S4, while maintaining the same structure as Mamba. The primary differences between the two S4 models lie in the application of multiplicative gating and the module order.\footnote{https://github.com/state-spaces/s4/blob/main/models/s4}

\vspace{-0.2cm}
\paragraph{Training.} We train each model by sampling a batch of random prompts at each training step and updating the model parameters using Adam optimizer~\citep{kingma2014adam}. We use a batch size of 64 and trained for 500,000 iterations (except for the vector MQAR task; see \cref{appendix:mqar_setup}).

\vspace{-0.2cm}
\paragraph{Evaluation.} We evaluate the model performance on in-context learning using task and data distributions $\mathcal{D}_{\mathcal{F}}$ and $\mathcal{D}_{\mathcal{X}}$ consistent to those during training. A function and a sequence of $N$ inputs are sampled from $\mathcal{D}_{\mathcal{F}}$ and $\mathcal{D}_{\mathcal{X}}$, respectively, to generate a test prompt $P_{\sf test}=(\bvec x_1, f(\bvec x_1), \ldots, \bvec x_N, f(\bvec x_N))$. We create 1,280 prompts and measure the empirical mean of Eq.~\eqref{eq:objective} across the prompts for in-context learning performance.

\new{Throughout our experiments, we keep the total number of parameters of models roughly the same for each configuration as explained in \cref{appendix:model_training}.}
To plot the model performance as the model capacity grows, we calculate the total floating point operations (FLOPs) used for training the model. The calculation for Transformer and Mamba can be found in ~\cref{appendix:flops}, which are based on \citep{Kaplan2020Scaling, Gu2023mamba}. 

Model configurations and training implementation details are provided in \Cref{appendix:setup}.

\subsection{In-context learning tasks}\label{sec:tasks}
We provide an overview of the ICL and related tasks investigated in this study. Some tasks are adapted from~\citep{Garg2022Transformers}, and we follow the settings outlined in their work. The tasks are summarized in \cref{tab:summary_of_tasks}.
\subsubsection{Learning regression}
For all regression tasks, in-context examples $\bvec x_i$ are sampled from the Gaussian distribution $\mathcal{N}(0, \bvec{I}_d)$, where $\bvec{I}_d$ is the $d\times d$ identity matrix. We use the squared error loss for model training.

\paragraph{Linear regression.}
We examine the class of linear functions $\mathcal{F}=\{f|f(\bvec x)=\bvec w^\top \bvec x, \bvec w\in\mathbb{R}^d\}$ where $\bvec w$ is sampled from the Gaussian distribution $\mathcal{N}(0, \bvec{I}_d)$. We set $d=20$.

\vspace{-0.2cm}
\paragraph{Sparse linear regression.}

The setting is identical to linear regression except that after sampling $\bvec w$ from $\mathcal{N}(0, \bvec{I}_d)$, $k$ coordinates are randomly retained and the rest are set to zero. We set $k=3$.

\vspace{-0.2cm}
\paragraph{Two-layer neural network.}
We consider the class of two-layer ReLU
neural networks $\mathcal{F}=\{f|f(\bvec x)=\bvec W^{(2)}\sigma\left(\bvec W^{(1)}\bvec x\right)\}$, where $\bvec W^{(2)}\in\mathbb{R}^{1\times h}, \bvec W^{(1)}\in\mathbb{R}^{h\times d}$, and $\sigma(\cdot)=\max(0, \cdot)$ is the ReLU function. Each element of the weight matrices is independently drawn from $\mathcal{N}(0,1)$. We use $d=20$ and $h=100$.

\vspace{-0.2cm}
\paragraph{Decision Tree}
We consider a full binary tree with a fixed depth and input $\bvec x\in\mathbb{R}^d$. Leaf node values are sampled from $\mathcal{N}(0, 1)$, and the rest are sampled uniformly at random from $\{1,...,d\}$, functioning as indices of $\bvec x$. 
At a given non-leaf node, we move to the right if $x[i] > 0$, where i is the sampled index, and otherwise move to the left.
$y$ is the leaf node value when the traversal terminates.

\subsubsection{Learning with outliers}

The problems that belong to this family adopt the basic setting of the standard linear regression task. With a fixed probability $p$, each pair of $({\bvec x}_i, f({\bvec x}_i))$ in the prompt is replaced with ``dummy'' vectors which are either out of the training distribution, or confounders designed to increase the complexity of the task. We test $p\in \{0.5, 0.9\}$ as the replacement probabilities for the tasks described below.
During training, we do not compute the loss for the replaced outliers. 

For evaluation, however, the locations of the dummy vectors were fixed to ensure that the models are evaluated on the same number of in-context examples across batches.
So we evaluate the loss at the 50th clean in-context example, where there is a clean example after every nine outliers for many-outlier ICL and after every one outlier for orthogonal-outlier ICL.

\paragraph{Orthogonal-outlier regression.}
    Each pair of $(\bvec{x}_i, f(\bvec{x}_i))$ is randomly replaced with $((a_x \bvec{u}+b_x \bvec{v})/(a_x^2+b_x^2), (a_y \bvec{u}+b_y \bvec{v})/(a_x^2+b_x^2))$, where $\bvec{u}, \bvec{v}\in \bvec{w}^{\perp}$. $(\bvec u, \bvec v) := (\bvec w_1-\text{proj}_{\bvec w}(\bvec w_1), \bvec w_2-\text{proj}_{\bvec w}(\bvec w_2))$ and $\bvec w_1$ and $\bvec w_2$ are sampled from $\mathcal{N}(0,\bvec I_d)$ and the coefficients $a_x,b_x, a_y, b_y$ are independently sampled from $\mathcal{N}(0,1)$. 

\vspace{-0.2cm}
\paragraph{Many-outlier regression.}
    In this setting, ${\bvec x}_i$ and $f({\bvec x}_i)$ are randomly replaced with a $d$-dimensional vector of ones $\{1\}^d$ and an one-hot vector $[1,0,\ldots, 0]$, respectively, with probability 90\%.
    Here, we test longer sequences of $N=512$, where only 10\% of the sequence yields useful information about the true target vector.

\subsubsection{Learning discrete functions}

\paragraph{Sparse parity.}
Following the setting from \citet{Bhattamishra2023TransformersLLMs}, we consider the class of functions $\mathcal{F} = \{f|f(\bvec x) = \prod_{j\in \mathcal{S}}\bvec x_i[j]\}$, where $\bvec x_i[j]$ denotes the $j$-th element of the vector $\bvec x_i$ and $\mathcal{S}$ is a subset of $\{1,\ldots, d\}$ with the size $k$. Each $\bvec x_i$ is sampled uniformly at random from $\{-1, 1\}^{d}$, and subset $\mathcal{S}$ of size $k$ is randomly sampled from the set $\{1,\ldots, d\}$. For this task, we train a model using the cross-entropy loss and evaluate the model using a binary indicator for accuracy, which assigns 1 to correct predictions and 0 to incorrect ones. 

\subsubsection{Learning Chain-of-Thought}

\paragraph{Chain-of-Thought-I/O.}

Following the setting from \citet{Li2023DissectingCoT}, we consider the class of two-layer ReLU neural networks $\mathcal{F}=\{f|f(\bvec x) = \bvec W^{(2)} \sigma\left(\bvec W^{(1)} \bvec x\right)\}$, where $\bvec W^{(2)} \in \mathbb{R}^{1\times h}, \bvec W^{(1)} \in \mathbb{R}^{h \times d}$, and $\sigma(\cdot)$ is the $\mathrm{ReLU}$ function. We set $d=10$ and $h=8$. We additionally interleave the intermediate hidden feature $\bvec s_i = \sigma\left(\bvec W^{(1)} \bvec{x}_i\right)$ in our input training sequence in a Chain-of-Thought (CoT) style. Given the input sequence $(\bvec x_1, \bvec s_1, f(\bvec x_1), \cdots, \bvec x_N, \bvec s_N, f(\bvec x_N), \bvec x_{\text{test}})$, the model is evaluated on the final output prediction $\hat{\bvec y}$ based on the input sequence and the intermediate layer prediction $\hat{\bvec s}_{\text{test}}$.

\subsubsection{Learning retrieval}

\paragraph{Vector multi-query associative recall.}

We test the model's ability to do multi-query associative recall (MQAR) \citep{Arora2023Zoology}. While MQAR is not an ICL task, model's ability to do associative recall (AR) is highly related to model's ability to learn in-context \citep{Olsson2022incontext}. To better measure the model's ability to retrieve information from context, we consider a variant of MQAR. The keys and the values are vectors, which can be interpreted as unique token embeddings. Specifically, the model is given a sequence of key-value pairs of vectors $\{\bvec{k}_1, \bvec{v}_1, ..., \bvec{k}_n, \bvec{v}_n\}$, where $\bvec{k}_i, \bvec{v}_i\in\mathcal{S}^{d-1}$ are sampled uniformly from the unit $d$-sphere. The query consists of sequence of vectors $\{\bvec{q}_1,...,\bvec{q}_m\}$. For each query $\bvec{q}_j$, there exists some $1\leq l\leq n$ such that $\bvec{q}_j = \bvec{k}_l$. The model must learn to output $\bvec{v}_l$ associated with the query $\bvec{q}_j$ for each of the queries, producing $m$ outputs total. We train models with mean squared error loss.

\subsubsection{Learning synthetic formal languages}

\new{Although not the main focus of our work, we conduct initial experiments using synthetic language benchmarks designed to assess in-context learning (ICL) capabilities within the language setting.
Given that real language ICL typically demands extensive datasets and computational resources, these synthetic datasets act as useful proxy for exploring language ICL. 
For detailed descriptions of their construction and evaluation, we direct readers to the respective publications.}

\textbf{GINC dataset}~\citep{Xie2021Explanation}.
Generative In-Context learning (GINC) dataset is a small-scale language dataset synthetically generated using a mixture of hidden markov models. Its pretraining dataset contains approximately 10 million tokens and each trained model is evaluated on 2500 test-time prompts containing 0 to 64 examples.
We train and test our models using a vocabulary size of 100.
We additionally train LSTMs for this dataset, as done in prior work.

\textbf{RegBench}~\citep{Akyurek2024ICLL}.
In-context Language Learning (ICLL) RegBench is a synthetic regular language benchmark created by randomly generating probabilistic finite automata (PFA) with uniform transition probabilities; 
multiple problem instances are produced that include samples from each PFA. 
The models are evaluated using a greedy-decoding accuracy metric, which assesses whether each next token predicted by the model is valid under the current regular language.

\section{In-context Learning Capabilities of Mamba}\label{sec:ssm_results}

In this section, we demonstrate that Mamba can be trained from scratch to perform various ICL tasks. 
Furthermore, we identify specific tasks in which one model performs better than others and vice versa, given the same amount of computation resources measured in terms of its total floating point operations (FLOPs) used in training. 

\begin{figure*}[ht]
  \begin{subfigure}[t]{\textwidth}
    \centering
    \includegraphics[width=0.58\textwidth]{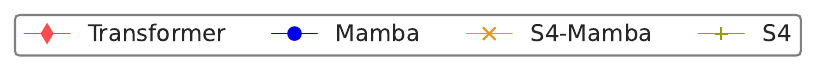}
    \centering
  \end{subfigure}

  \begin{subfigure}[b]{0.49\textwidth}
    \centering
    \includegraphics[width=\textwidth]{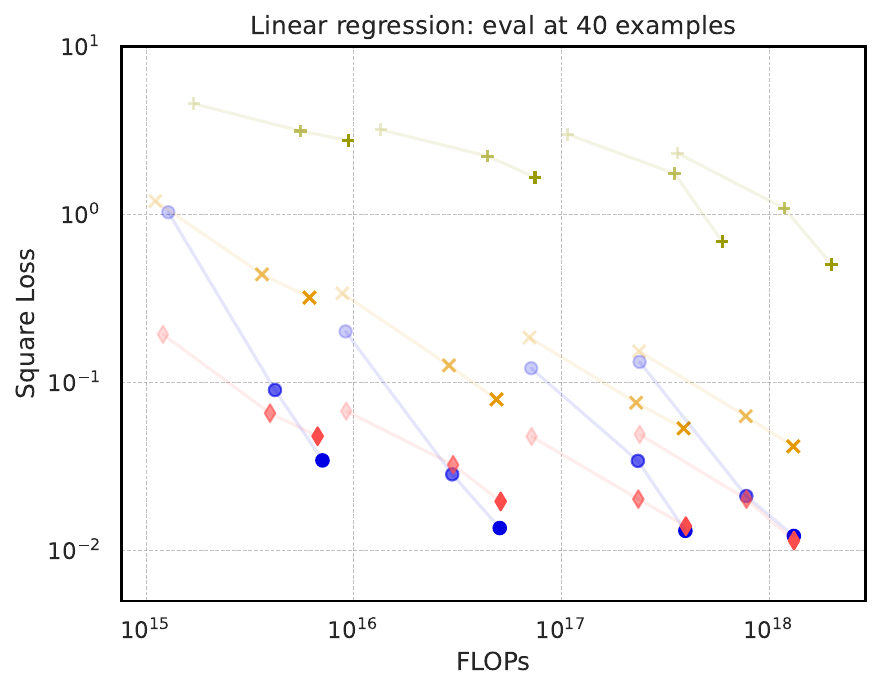}
  \end{subfigure}%
  \begin{subfigure}[b]{0.49\textwidth}
    \centering
    \includegraphics[width=\textwidth]{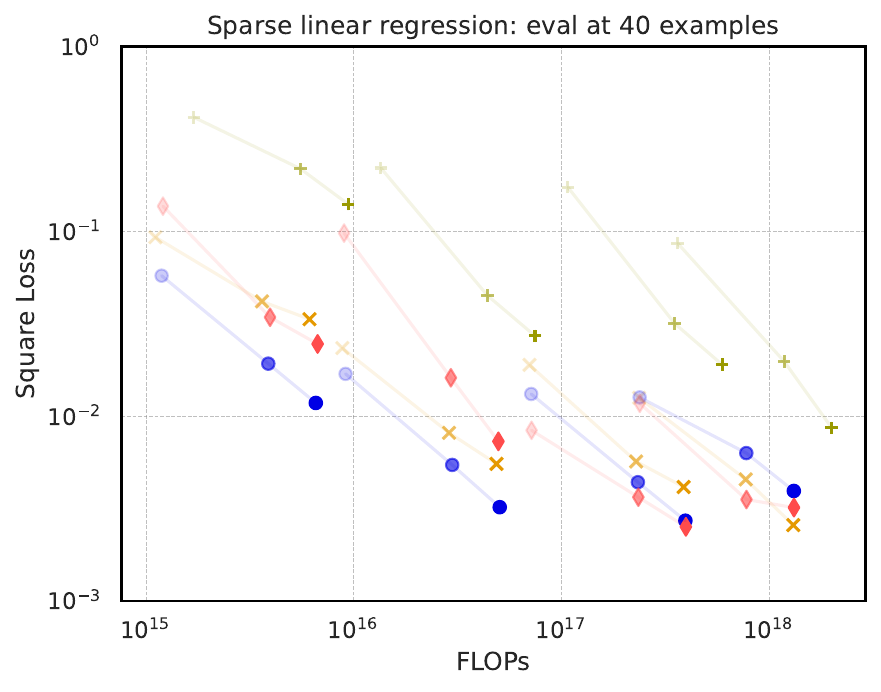}
  \end{subfigure}

  \vspace{-4mm}
  \begin{subfigure}[t]{\textwidth}
    \centering
    \label{row:2}
  \end{subfigure}
  \vspace{-4mm}

  \begin{subfigure}[b]{0.48\textwidth}
    \centering
    \includegraphics[width=\textwidth]{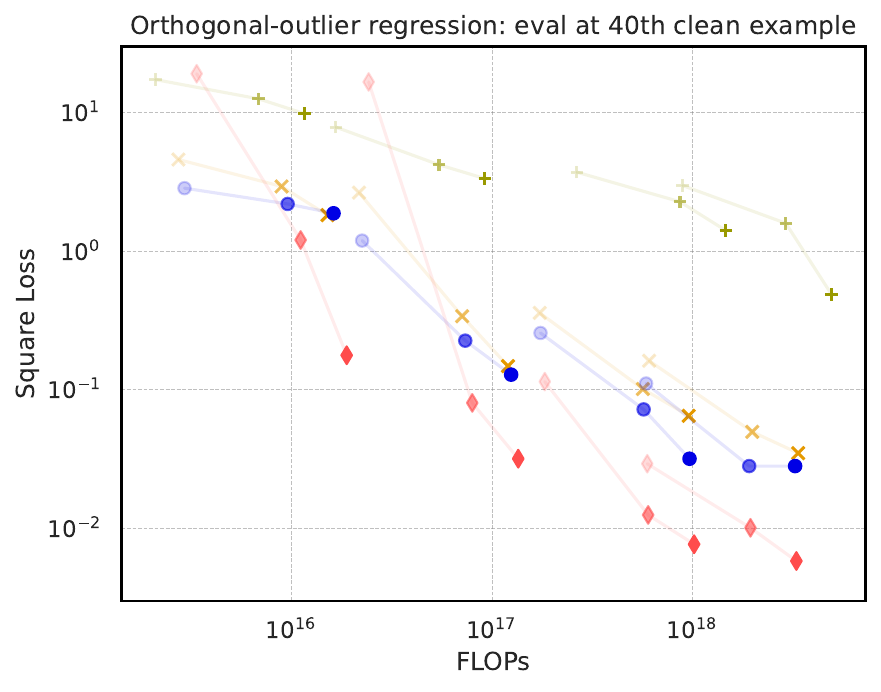}
  \end{subfigure}
  \hspace{1mm}
  \begin{subfigure}[b]{0.5\textwidth}
    \centering
    \includegraphics[width=\textwidth]{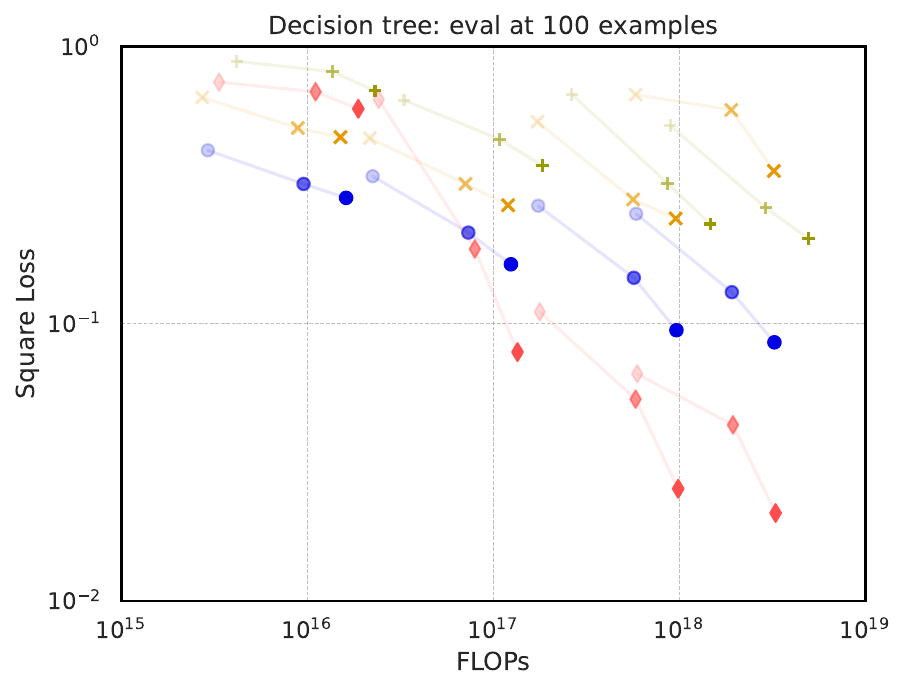}
  \end{subfigure}%
  
  \vspace{-4mm}
  \begin{subfigure}[t]{\textwidth}
    \centering
  \end{subfigure}
  \vspace{-4mm}

  \begin{subfigure}[b]{0.5\textwidth}
    \centering
    \includegraphics[width=\textwidth]{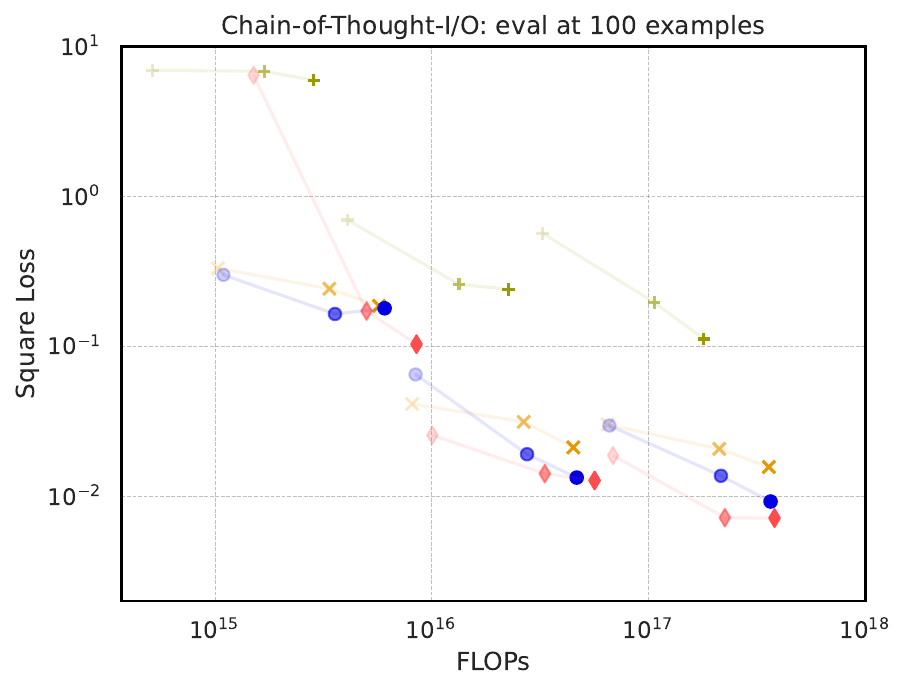}
  \end{subfigure}%
  \begin{subfigure}[b]{0.51\textwidth}
    \centering
   \includegraphics[width=\textwidth]{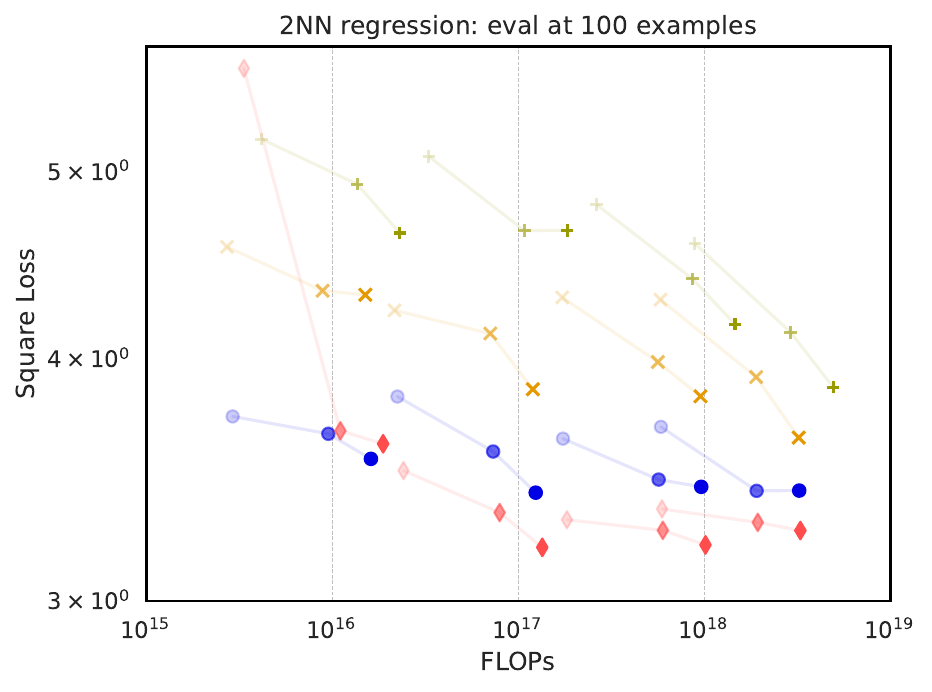}
  \end{subfigure}
  \caption{Model performance on our suite of ICL tasks for Transformer, Mamba, S4-Mamba, and S4 where each color represents a different architecture. For each architecture, the best performing model given the same amount of FLOPs is plottet (see \Cref{appendix:setup} for details on model configurations). Transparent points indicate earlier stages of training; plotted models are trained in between \{100k, 300k, 500k\} iterations. The descriptions of tasks can be found in \Cref{sec:tasks}.}
  \label{fig:scaling_ssm}
\end{figure*}

\subsection{Mamba {\it can} in-context learn!}

\begin{highlight}
\paragraph{Finding 1:} 
\emph{Mamba outperforms its simpler counterparts, while performing as well as Transformer on a range of ICL tasks.}
\end{highlight}

In \cref{fig:scaling_ssm}, Mamba consistently outperforms its simpler counterparts S4-Mamba and S4. For linear regression, the gap between Mamba and S4-Mamba is much smaller than that between S4-Mamba and S4.
As the only difference between Mamba and S4-Mamba is the input-dependent selection mechanism,
appropriate gating and stacking of MLPs (\textit{i.e.}, difference between S4-Mamba and S4) seem to be more significant for such tasks.
In comparison, the input-dependence of Mamba makes meaningful progress for more complex tasks such as 2NN regression and learning decision trees.

Mamba can also perform on par with Transformer even as the total FLOPs scale up.
This is surprising given that Transformer and attention have been the focus of many previous works for its unique ICL capability.
Moreover, Mamba tends to perform better in smaller parameter settings when controlling for equal depth, \emph{i.e.}, keeping the number of attention, MLP, and Mamba blocks equivalent.

\subsection{Performance gaps in more complex ICL tasks}

\begin{figure*}[ht]
  \begin{subfigure}[t]{\textwidth}
    \centering
    \includegraphics[width=0.94\textwidth]{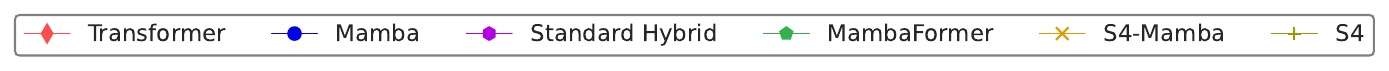}
    \centering
  \end{subfigure}

  \begin{subfigure}[b]{0.5\textwidth}
    \centering
    \includegraphics[width=\textwidth]{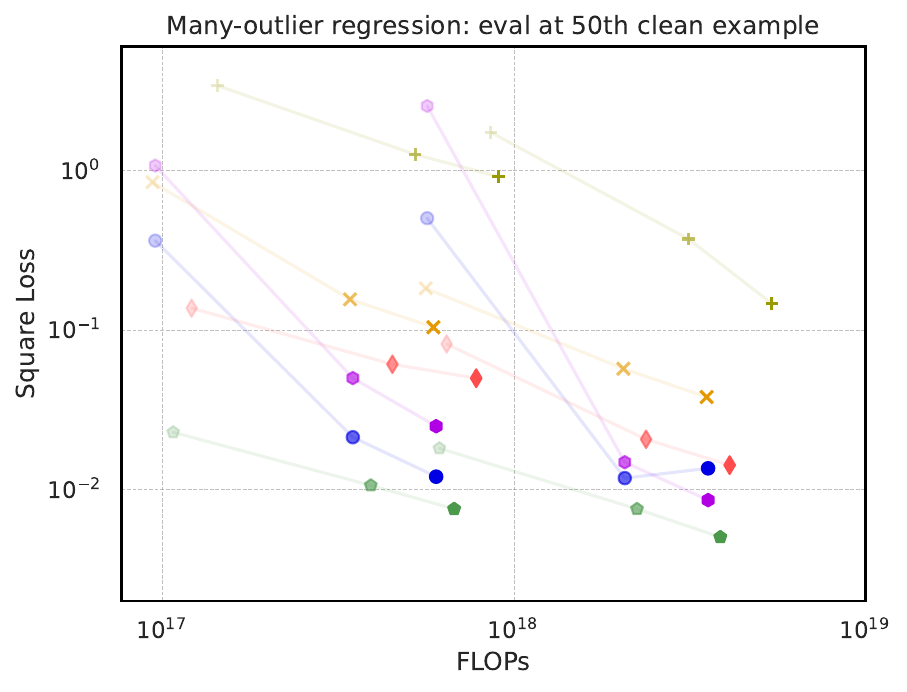}
  \end{subfigure}%
  \begin{subfigure}[b]{0.5\textwidth}
    \centering
    \includegraphics[width=\textwidth]{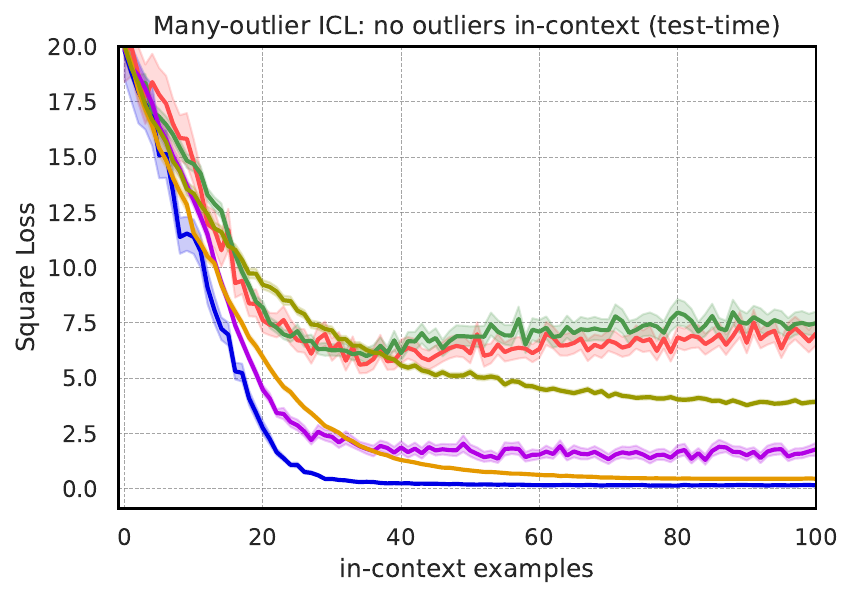}
  \end{subfigure}

  \vspace{-4mm}
  \begin{subfigure}[t]{\textwidth}
    \centering
  \end{subfigure}

  \begin{subfigure}[b]{0.49\textwidth}
    \centering
    \includegraphics[width=\textwidth]{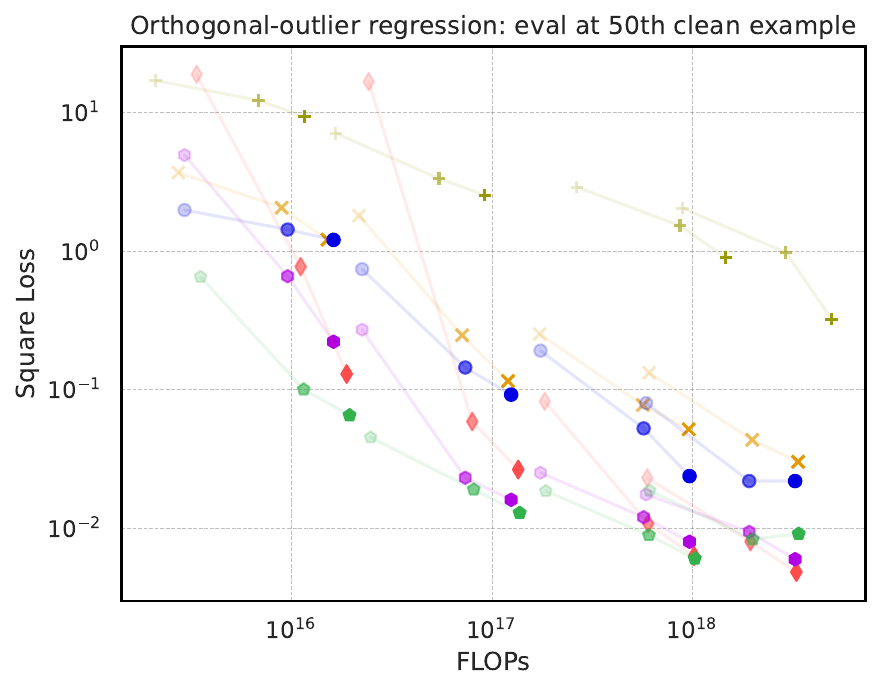}
  \end{subfigure}
  \hspace{1mm}
  \begin{subfigure}[b]{0.49\textwidth}
    \centering
    \includegraphics[width=\textwidth]{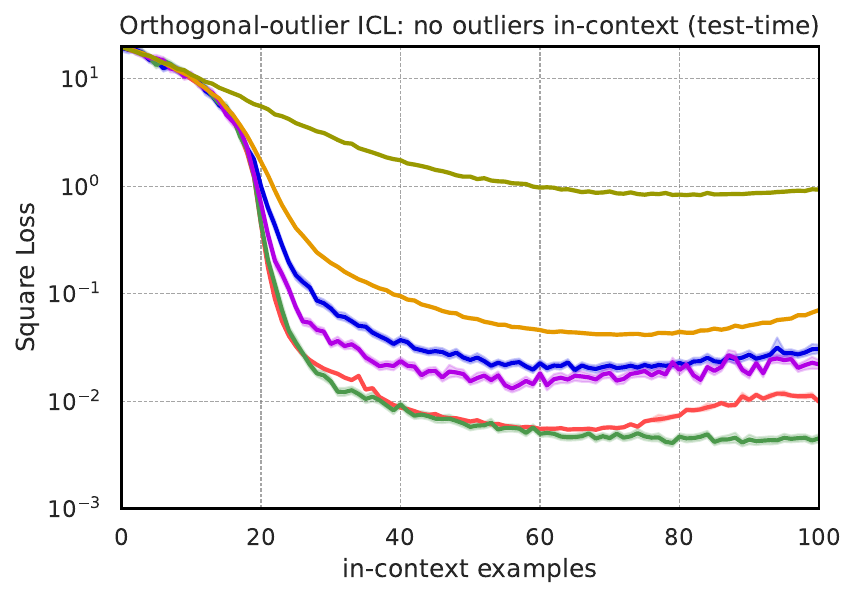}
  \end{subfigure}%
  \caption{\textbf{(Left)} Performance of various architectures on two robust linear regression tasks.  More transparent points indicate earlier stages of training; plotted models are trained in between \{100k, 300k, 500k\} iterations. \textbf{(Right)} Out-of-distribution performances when models do not see outliers during test-time, \textit{i.e.}, standard linear regression. Task descriptions can be found in \cref{tab:summary_of_tasks}. 
  \vanilla and \variant are hybrid models of Transformer and Mamba defined in~\cref{sec:hybrid}. }
  \label{fig:ortho_result}
\end{figure*}

\begin{figure*}[ht]
  \centering
  \begin{subfigure}[b]{0.48\textwidth}
      \includegraphics[width=\textwidth]{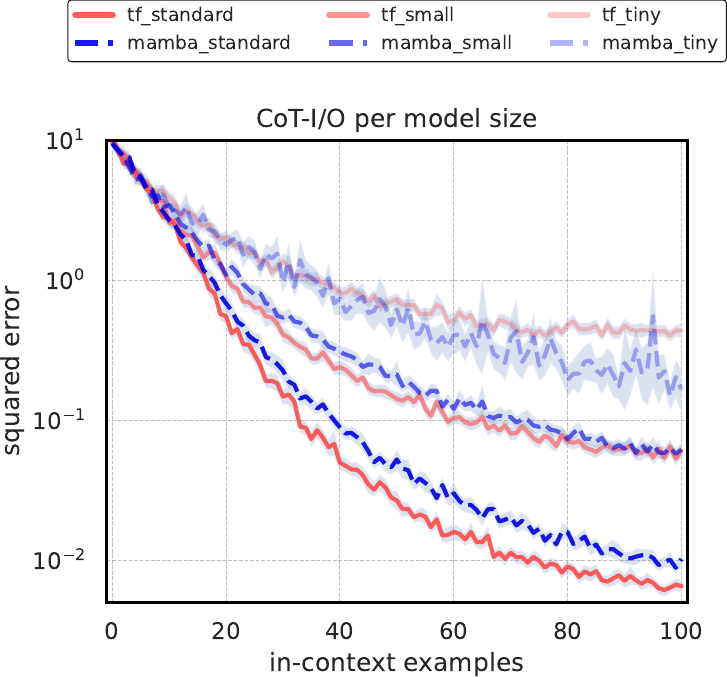}
  \end{subfigure}
    \begin{subfigure}[b]{0.48\textwidth}
      \includegraphics[width=0.98\textwidth]{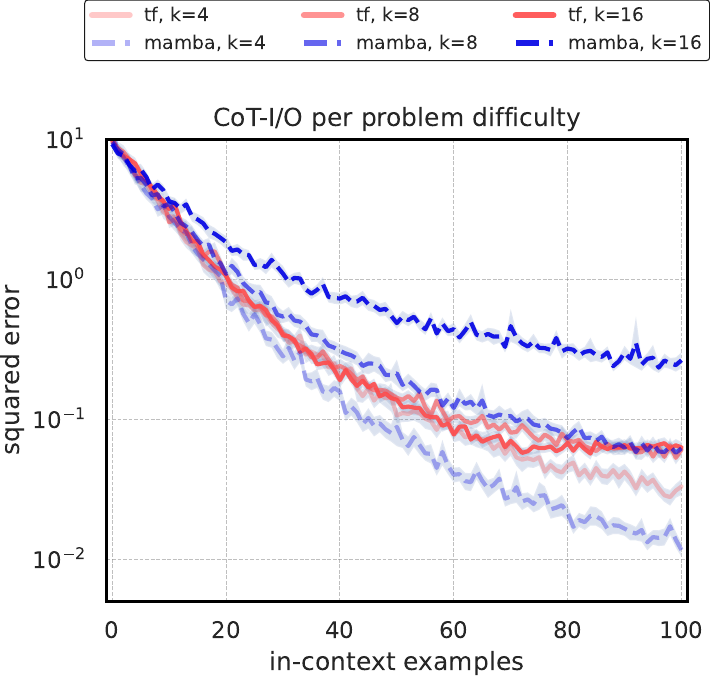}
  \end{subfigure}
  \caption{Performance of Transformer and Mamba models on the Chain-of-Thought-I/O task. Experiments on varying the model size (left) and varying the hidden dimension (right). Model configurations can be found in \Cref{appendix:setup} \Cref{tab:cot_config}.}
  \label{fig:CoT-IO_result}
\end{figure*}

We also consider a family of more complex ICL tasks, namely learning decision tree, sparse parity, outlier-robust regression (\Cref{fig:ortho_result}) and Chain-of-Thought (\Cref{fig:CoT-IO_result}). 
We elaborate on the performances of each model on each task in our findings below.

\begin{highlight}
\paragraph{Finding 2:} 
\emph{For outlier-robust regression, Mamba outperforms Transformer in ignoring prevalent fixed outliers, while Transformer is better when the outliers are not fixed.}
\end{highlight}
\vspace{2mm}

Orthogonal-outlier regression and many-outlier regression, like other outlier tasks, focus on the model's ability to learn to ignore dummy vectors, either by the fact that the $\bvec x_i \in \bvec w^{\perp}$, or by the fact that $\bvec y_i$ is a vector instead of a zero-padded scalar value. This explicitly requires the models to look at previous input sequences and discover the properties that distinguish the dummy vectors from training examples while learning the class of functions the training prompt represents.

For orthogonal-outlier regression task with a relatively short sequence length of 101, \new{Mamba does not perform as well as Transformer given the same total FLOPs, though its learns significantly better than S4} (\cref{fig:ortho_result}).
However, for many-outlier regression where we train on a sequence length of 512 and 90\% all-ones replacement, \new{Mamba outperforms Transformers, especially in terms of its out-of-distribution (OOD) accuracy where we evaluate each model on clean sequences with no outliers at all.
Recurrent models, such as S4 and Mamba, seem to generalize well in such OOD regime when the data is contaminated with many identical outlier vectors.}
This is also in line with what \citet{Gu2023mamba} reports: Mamba fares better in retrieval tasks of long sequence lengths \new{with a single retrieval key}.
These results indicate that Mamba has no significant issue with filtering out unnecessary information, while retaining the ability to learn linear regression in-context.

\vspace{2mm}
\begin{highlight}
\paragraph{Finding 3:} 
\emph{For Chain-of-Thought-I/O, Mamba shows comparable performance to Transformer.}
\end{highlight}
\vspace{2mm}

\Cref{fig:scaling_ssm} and \Cref{fig:CoT-IO_result} shows that Mamba models are capable of in-context learning in a chain-of-thought manner, showing comparable performance to Transformer models across the tested configurations. In smaller model configurations, Mamba models exhibit superior performance compared to Transformer models. However, as the model size increases, Transformer models begin to surpass Mamba models. The performance of Transformer models remains relatively stable across different problem sizes, while Mamba models' performance is significantly influenced by the size of the hidden layer. Specifically, Mamba models excel over Transformer models at smaller problem sizes (\textit{i.e.}, smaller hidden dimensions), but their advantage diminishes as the problem size expands.

\subsection{Challenges in parity and retrieval}
\label{sec:mqar_result}

We run vector MQAR on two settings: (1) $32$ key-value pairs with $16$ queries and (2) $32$ key-value pairs with $4$ queries.

\begin{table}[ht]
\centering
    \begin{tabular}{lcccc}
    \toprule
    Number of queries & \multicolumn{2}{c}{4} & \multicolumn{2}{c}{16} \\
    \cmidrule(lr){2-3} \cmidrule(lr){4-5}
    Embedding dimension &  64 & 128 & 64 & 128 \\ \midrule
    Mamba   & 8.64e-1 & 1.64e-1  & 7.23e-1 & 1.50e-1 \\ 
    Transformer without PE  & 1.14e-3 & 8.66e-5 &  7.61e-5 & 5.55e-5 \\
    Transformer with PE  & \textbf{5.17e-6} & \textbf{8.76e-7} & 3.99e-5 & \textbf{2.46e-7}\\ 
    \variant   & 7.30e-6 & 3.37e-6 & \textbf{1.03e-5} & 3.79e-7 \\
    6 Mamba Blocks + 1 \vanilla   & 1,99e-2 & 1.37e-2 & 1.54e-3 & 5.86e-5\\ 
    \bottomrule \hfill
    \end{tabular}
    \caption{Test loss (mean squared error) on vector MQAR and respective model configurations. We test Transformers with and without Positional Encoding (PE). All models have $4$ layers with roughly the same number of parameters. We consider the ``$6$ Mamba Blocks + $1$ \vanilla'' model as $4$ layers since one Mamba layer consists of two Mamba blocks as described in \cref{fig:model_arch}.}
   \label{tab:mqar_res1}
\label{tab:mqar}
\end{table}

\begin{highlight}
\paragraph{Finding 4:} 
\emph{Mamba struggles to retrieve vectors within its context in vector MQAR, a task Transformer can easily perform.}
\end{highlight}
\vspace{2mm}

From \cref{tab:mqar}, we can see that Mamba struggles to accurately retrieve the vectors as the mean squared error for retrieving normalized vectors are greater than $0.1$ in all cases. 
Since SSMs are limited by their hidden state dimension in carrying information to predict the next token, they would eventually be overwhelmed if the number of key-value pairs within the context (not queries) increases substantially.

Note that the models trained with $16$ queries have lower test loss than models trained with $4$ queries. We conjecture that for a single sequence of data that represents an MQAR task, each $(\bvec q, \bvec v)$ pair can be thought of as a training sample. Hence a sequence with $16$ queries contains more training samples than that of a sequence with $4$ queries. This also shows that having more queries does not necessarily make the task harder.
Notably, our setting is more challenging than token-based MQAR, as we sample new random vectors each batch. Similar findings on retrieval were observed in~\citet{Arora2023Zoology}.

\vspace{2mm}
\begin{highlight}
\paragraph{Finding 5:} 
\emph{Mamba can in-context learn sparse parity, a task Transformer cannot perform.}
\end{highlight}
\vspace{2mm}

\begin{figure}[ht]
  \centering
  \begin{subfigure}[b]{0.7\textwidth}
      \includegraphics[width=\textwidth]{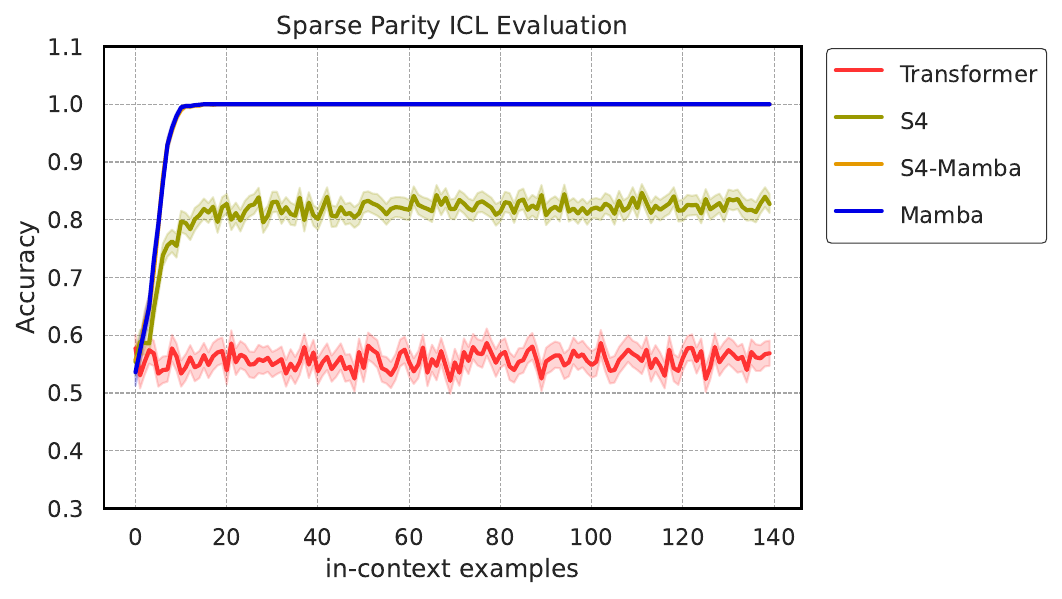}
  \end{subfigure}
  \caption{Although Transformer struggles to learn the task, Mamba and S4-Mamba can learn sparse parity of $d=10$ and $k=2$ (S4-Mamba accuracy plot is hidden behind that of Mamba). Each model is trained with an embedding dimension of 256 and depth of 12 layers (approximately 10M parameters) up to 500,000 iterations. Transformer struggles to learn even up to an embedding dimension of 768 and 24 layers and 1M iterations.}
  \label{fig:sparse_parity}
\end{figure}

While Mamba fails on simple retrieval tasks such as MQAR, 
the tables turn for the task of learning sparse parity (\cref{fig:sparse_parity}).
Transformer fails to do better than random guessing, in line with evidence from prior work~\citep{Bhattamishra2020FormalLanguages, Bhattamishra2023TransformersLLMs, Hahn2020Theoretical}.
We confirm this is the case for Transformer sizes of embedding dimensions up to 768 and up to 24 layers when trained for at most 1 million iterations.
However, Mamba succeeds in this task with ease, solving sparse parity for $(d,k) = (10,2)$ with a network as small as 2 layers. 

Even more surprisingly, S4-Mamba is able to solve parity as well, showing comparable performance to that of Mamba; this indicates that proper convolution or gating may be more important than input-dependent selection for learning parity. \new{Given that only Transformer cannot perform better than random, sequential computations of recurrent models seem more advantageous for learning parity.}
Finally, our result hints at that the initial (causal) convolution that Mamba provides before the attention layer may be crucial to solving parities, a similar phenomenon observed for Vision Transformers in computer vision tasks~\citep{Yu2022Metaformer}.

Any algorithm for learning parities requires either a super-linear memory of $\omega (d)$ or a super-polynomial number of samples in $d$~\citep{Raz2016Parity, Kol2017Parity}.
While Transformer is known to have better memory than Mamba due to its quadratic attention mechanism, our result on learning sparse parities brings forth the question on how different architectures may utilize its memory differently in terms of function approximation.
We leave the theoretical and empirical question of which architectural component allows for learning parities as an avenue for further study.

\section{The Advantage of Hybrid Architectures for In-context Learning}
\label{sec:hybrid}

\begin{figure*}[hb]
  \centering\includegraphics[width=0.96\textwidth]{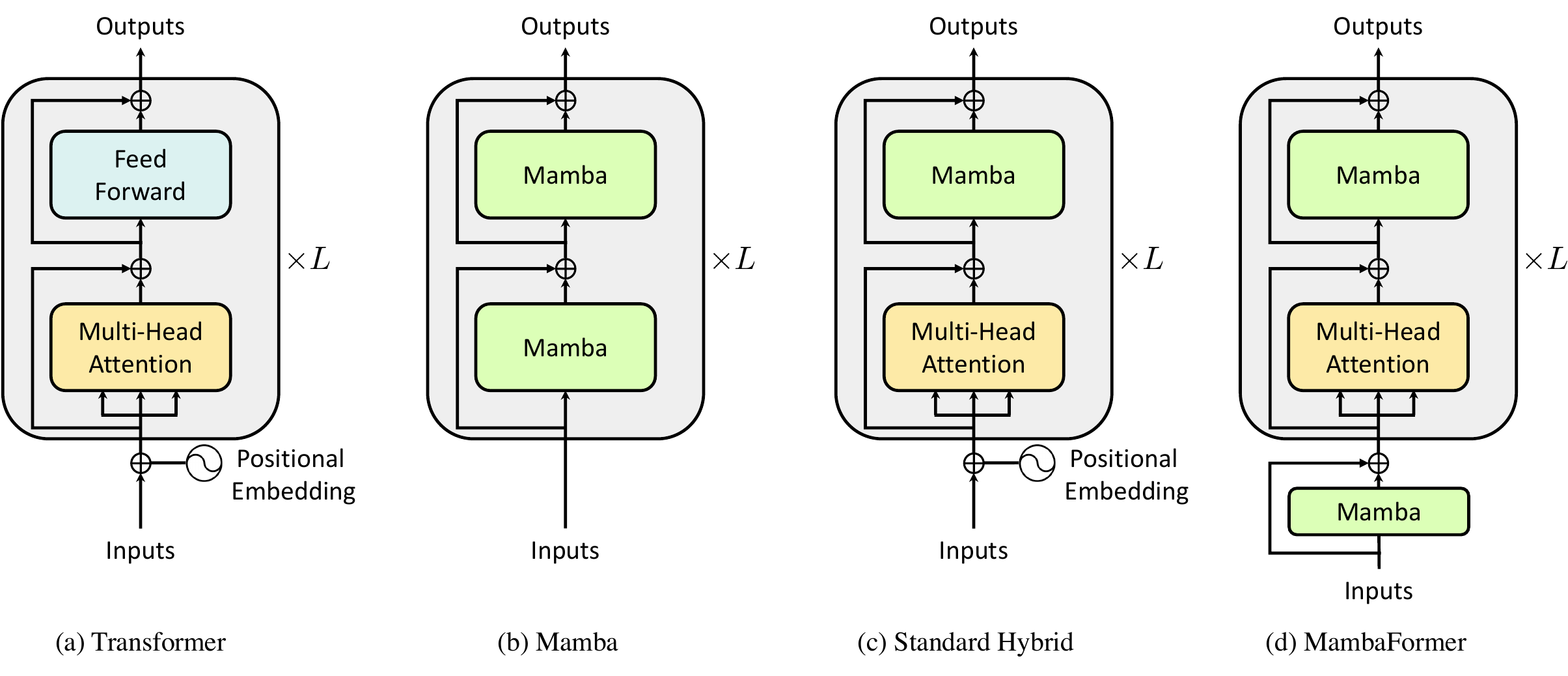}
  \caption{Model architectures. (a) and (b) are the standard Transformer and Mamba architectures. (c) denotes the hybrid architecture of Mamba and Attention blocks, following the design proposed in \citet{Gu2023mamba}. (d) depicts the proposed architecture, namely \variant, which replaces the Positional Encoding with a Mamba block. We denote 2 blocks of either Mamba, multi-head Attention, or a feed forward network as 1 layer.}
  \label{fig:model_arch}
\end{figure*}

In the previous section, we have observed that Transformers perform better than SSMs in certain tasks such as learning decision trees or retrieval, while SSMs excel in others, such as learning sparse parities or heavy-outlier linear regression, possibly due to its recurrent nature. However, can we achieve the best of both worlds without sacrificing performance in our suite of ICL tasks?

We answer this in the affirmative; in this section, we investigate two hybrid architectures that combine Transformer and Mamba, namely \vanilla and \variant as illustrated in \cref{fig:model_arch}. \vanilla is the architecture of interleaving MHA and Mamba by replacing the MLP block with Mamba. \variant is nearly identical to \vanilla but with an additional Mamba block as its initial layer. This removes the need of initial positional encoding as a Mamba block's recurrent nature encodes positional information.

Although many works have found that interleaving multi-head attention and LTI SSMs beneficial~\citep{Zuo2022Efficient, Mehta2022Long, Pilault2023Block}, interestingly \citet{Gu2023mamba} have not found significant benefits of interleaving.

In the following results, however, we show that we can indeed reach competitive performance in our suite of ICL tasks by interleaving Attention and Mamba blocks. \variant achieves comparable performance to that of Transformer or Mamba, while excelling in both sparse parity and retrieval, tasks unsolvable by Transformer and Mamba, respectively. We discover that the key ingredient is having Mamba as the first layer.

\subsection{Simultaneously learning parities and retrieval}

\begin{highlight}
\paragraph{Finding 6:} 
\emph{\variant can in-context learn sparse parity; moreover, having the initial layer as a Mamba block is significantly effective.}
\end{highlight}

\begin{figure}[ht]
  \centering
  \begin{subfigure}[b]{0.75\textwidth}
      \includegraphics[width=\textwidth]{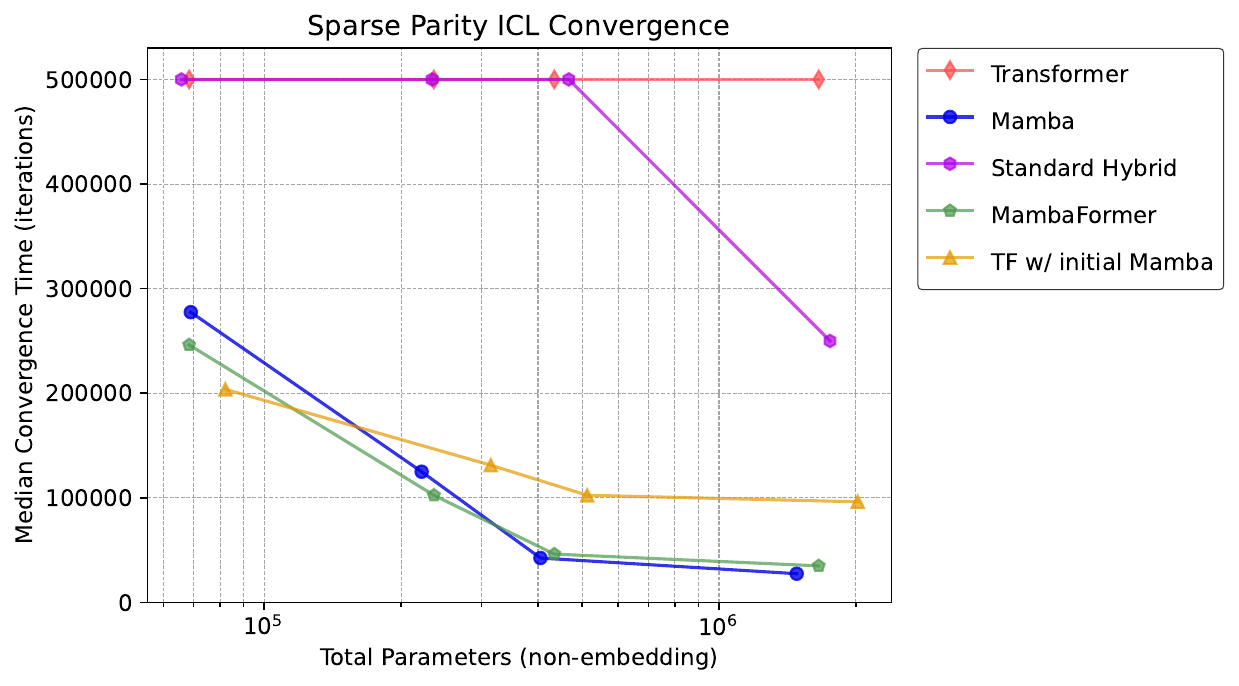}
  \end{subfigure}
  \caption{Median convergence time of learning parity over 5 random seeds for max. 500k iterations, where 500k convergence time signifies failed learning. Having the initial layer as Mamba is essential for efficiently learning parities. Tested model configurations are specified in~\cref{appendix:setup}.}
  \label{fig:parity_converge}
\end{figure}

As highlighted in \citet{Bhattamishra2023TransformersLLMs, Barak2022Hidden}, learning sparse parity in-context seems to be difficult for Transformer and some SSMs like Hyena.
Yet interestingly, as seen in \cref{fig:parity_converge}, \variant successfully learns parity as quickly as Mamba in terms of sample complexity.
While the \vanilla model is also capable, it exhibits much worse sample efficiency.

We perform an ablation study by equipping Transformer with an initial Mamba block without any positional encoding. Although this variant Transformer only has fewer Mamba blocks than \vanilla,
it solves parity almost as efficiently as Mamba.
Not only does this show us that the order of layers in interleaving matters as shown in \citet{press2022measuring}, but also that Mamba can complement Transformer without hurting performance in ICL.
This result brings up intriguing differences between the function learning capabilities of Attention and Mamba; we leave this question up for further study.

\begin{highlight}
\paragraph{Finding 7:} 
\emph{\variant can perform retrieval as well as Transformer, closing the performance gap between Mamba and Transformer.}
\end{highlight}

The gap between Mamba and Transformer in vector MQAR is likely 
due to the fact that Mamba (as an SSM) compresses context into smaller states when generating output, while the Attention mechanism in Transformer does not compress the context. 
The amount of context SSMs store at each state depends on the dimension of hidden state as the hidden states capture the important information in the context. 
In contrast, attention leverages all tokens in its input context, allowing Transformers and hybrid models to conveniently retrieve corresponding key-value pairs through pairwise computations. On the other hand, SSMs would eventually be overwhelmed if the number of key-value pairs increases substantially. 

To close the gap in the vector MQAR task between Mamba and Transformer without sacrificing efficiency too much, we add one attention layer within the Mamba layers. In particular, in a Mamba model of $4$ layers ($8$ Mamba blocks stacked homogeneously), we replace the middle two blocks with \vanilla (w/o positional embedding). As shown in \cref{tab:mqar}, Mamba model gains a significant improvement in vector MQAR by having one \vanilla. We further test \variant on the same task and find that \variant almost entirely closes the gap to transformer in vector MQAR task.

\subsection{All-in-one ICL performance}

\begin{figure*}[ht]
  \begin{subfigure}[t]{\textwidth}
    \centering
    \includegraphics[width=0.75\textwidth]{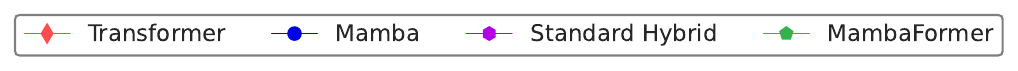}
    \centering
  \end{subfigure}

  \begin{subfigure}[b]{0.49\textwidth}
    \centering
    \includegraphics[width=\textwidth]{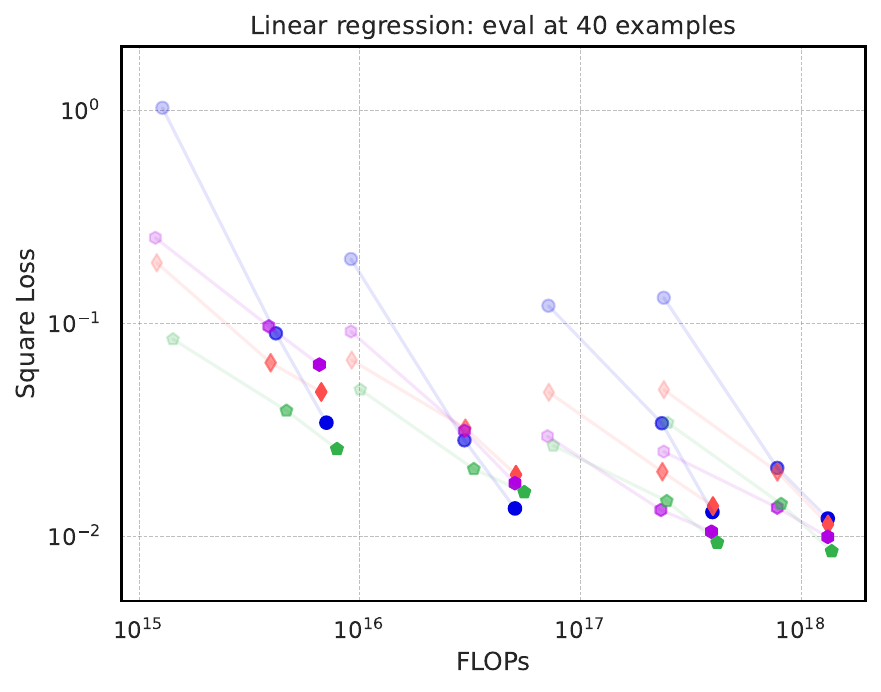}
  \end{subfigure}%
  \begin{subfigure}[b]{0.49\textwidth}
    \centering
    \includegraphics[width=\textwidth]{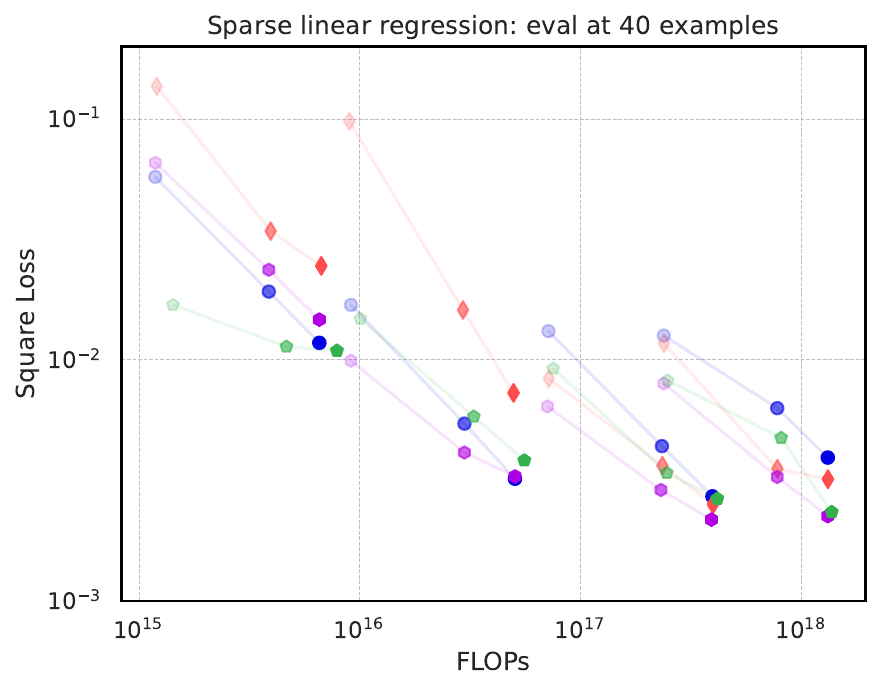}
  \end{subfigure}

  \vspace{-4mm}
  \begin{subfigure}[t]{\textwidth}
    \centering
  \end{subfigure}
  \vspace{-4mm}

  \begin{subfigure}[b]{0.48\textwidth}
    \centering
    \includegraphics[width=\textwidth]{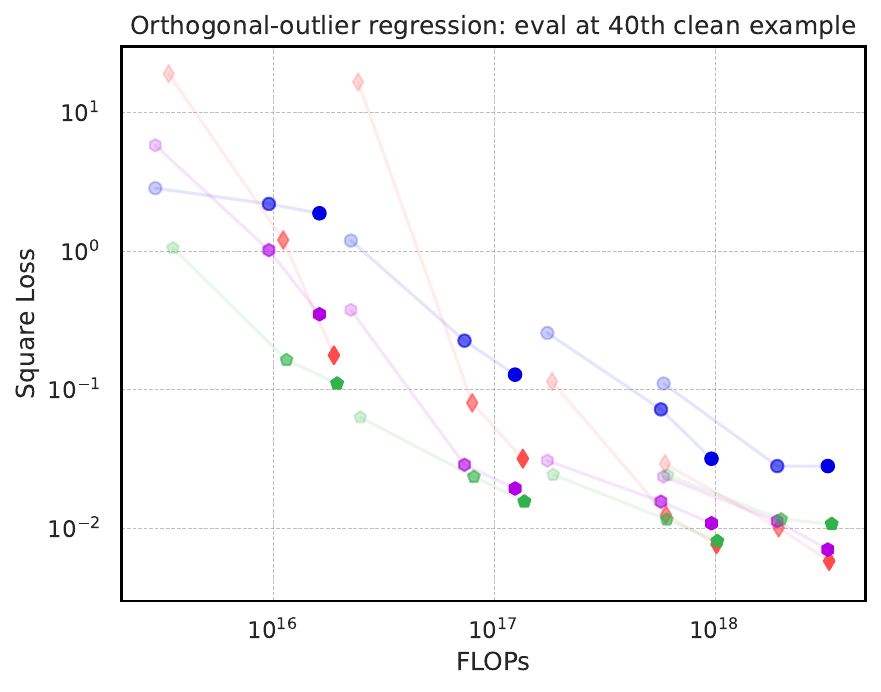}
  \end{subfigure}
  \hspace{1mm}
  \begin{subfigure}[b]{0.5\textwidth}
    \centering
    \includegraphics[width=\textwidth]{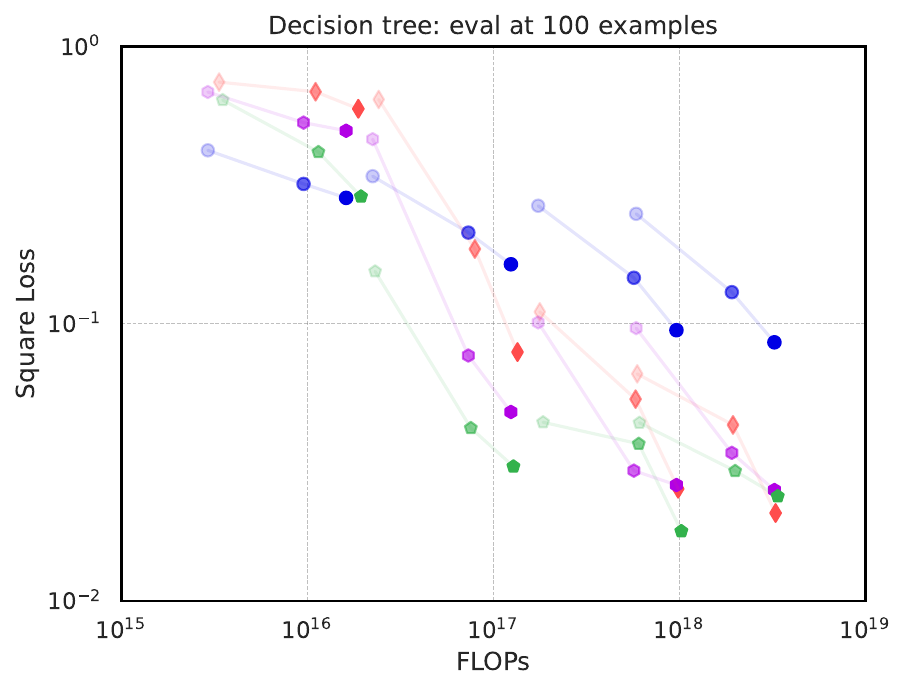}
  \end{subfigure}%
  
  \vspace{-4mm}
  \begin{subfigure}[t]{\textwidth}
    \centering
  \end{subfigure}
  \vspace{-4mm}

  \begin{subfigure}[b]{0.5\textwidth}
    \centering
    \includegraphics[width=\textwidth]{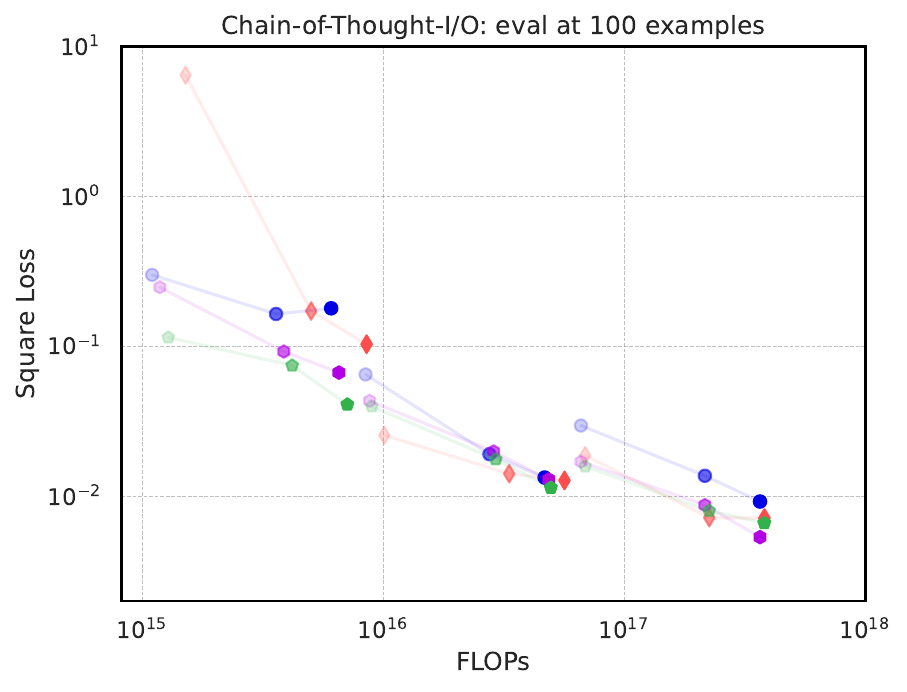}
  \end{subfigure}%
  \begin{subfigure}[b]{0.51\textwidth}
    \centering
   \includegraphics[width=\textwidth]{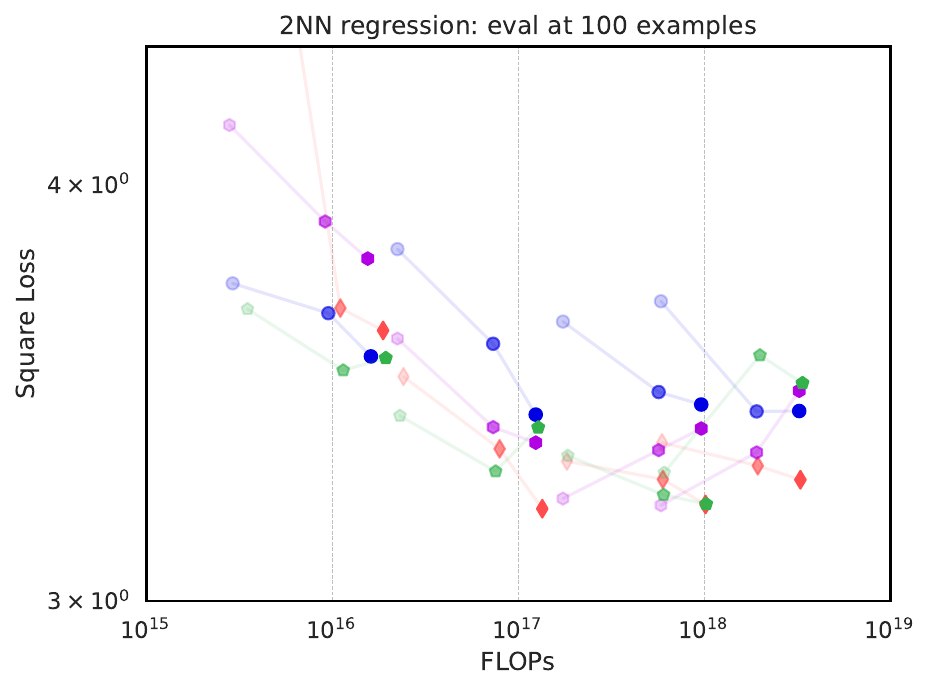}
  \end{subfigure}
  \caption{A suite of ICL tasks ran for Transformer, Mamba, and hybrid architectures where each color represents a different architecture. More transparent points indicate earlier stages of training; plotted models are trained \{100k, 300k, 500k\} iterations.
  \vanilla and \variant are hybrid models of Transformer and Mamba defined in~\cref{sec:hybrid}.
  }
  \label{fig:scaling}
\end{figure*}

\begin{highlight}
\paragraph{Finding 8:} 
\emph{Both hybrid models perform as well as Transformer and Mamba in our suite of ICL tasks (or even better sometimes).}
\end{highlight}

While \variant succeeds in two tasks that were deemed difficult for either Mamba or Transformer, it also performs equally well as Transformer and Mamba do in the rest the ICL tasks.
In \Cref{fig:scaling}, we see that \variant and \vanilla both learn decision trees as well as Transformer does and better than Mamba, even at larger parameter sizes.

More surprisingly, \variant efficiently learns linear regression more robustly even in the presence of  noisy data in Many-outlier regression and Orthogonal-outlier regression (see \Cref{fig:ortho_result}). In particular, a small \variant trained on 100k iterations ($< 10^{17}$ FLOPs) performs as well as models trained with nearly 5 times the number of FLOPs (\Cref{fig:ortho_result} left). 

\new{When evaluated with no outliers during test-time, \variant resembles Transformer and \vanilla resembles Mamba in terms of its out-of-distribution performance, where Mamba easily learns linear regression when there is only one outlier vector (\Cref{fig:ortho_result} top right) while Transformer learns better when there is a subspace of outlier vectors (\Cref{fig:ortho_result} bottom right).}

In conclusion, we find the best of both worlds within our diverse array of ICL tasks; a hybrid architecture that can solve as difficult problems as retrieval and parity, while performing on par with Transformer and Mamba in other ICL tasks.
Given our results, it will be interesting to see how hybrid architectures perform in other kinds of ICL tasks, especially in-context language benchmarks such as \citet{Xie2021Explanation, Hahn2023theory, Akyurek2024ICLL}.
\new{In turn, we explore formal language ICL capabilities in the following subsection.}

\subsection{In-context learning formal languages}

Given the empirical strength in hybrid models, this subsection analyzes their performance on synthetic formal language benchmarks, namely GINC and ICLL RegBench.
We use these benchmarks as a proxy to measure language ICL capabilities.

\begin{table}[h]
\centering
\begin{tabular}{@{}lcccc@{}}
\toprule
\textbf{GINC} & \textbf{Parameters} & \textbf{Train PPL} $(\downarrow)$ & \textbf{Valid PPL} $(\downarrow)$ & \textbf{ICL acc.} $(\uparrow)$ \\
\hline
LSTM            & 29M  & \textbf{3.53}  & \textbf{3.71} & \textbf{96.4} $\pm$ 0.6   \\
Transformer     & 86M  & 4.06  & 4.14  & 84.2 $\pm$ 5.1  \\
Mamba           & 90M  & 4.30  & 4.57   & 87.1 $\pm$ 7.8  \\
\variant     & 77M  & 4.22  & 4.77   & 79.6 $\pm$ 3.8 \\
\vanilla     & 74M  & 4.18  & 4.65   & 85.0 $\pm$ 3.1 \\
\bottomrule
\hfill
\end{tabular}
\caption{GINC data has a vocab size of 100 and the ICL accuracy is evaluated at 64 examples, where each example has length 10. Each model is trained with embedding size 768 and 12 layers, other than LSTM, which used embedding size 768, hidden layer size 768, and 6 layers. We include 90\% confidence intervals for ICL accuracy. We follow the same training recipes as \citet{Xie2021Explanation}.}
\label{tab:ginc}
\end{table}

\begin{table}[htbp]
\centering
\begin{tabular}{@{}lccc@{}}
\toprule
\textbf{RegBench} (trained 15 epochs) & \textbf{Train PPL} $(\downarrow)$ & \textbf{Valid PPL} $(\downarrow)$ & \textbf{Acc.} $(\uparrow)$ \\
\midrule
LSTM            & 6.20  & 6.39 & 51.0   \\
Transformer     & 4.20  & 4.17  & 92.6*  \\
Mamba           & 5.59  & 5.69   & 69.4  \\
\variant     & \textbf{1.01}  & \textbf{1.01}   & 99.8 \\
\vanilla     & \textbf{1.01}  & \textbf{1.01}   & \textbf{99.9} \\
\bottomrule
\end{tabular}

\vspace{5mm} 

\begin{tabular}{@{}lccc@{}}
\toprule
\textbf{RegBench} (trained 120 epochs) & \textbf{Train PPL} $(\downarrow)$ & \textbf{Valid PPL} $(\downarrow)$ & \textbf{Acc.} $(\uparrow)$ \\
\midrule
LSTM            & 3.33  & 4.37 & 73.5  \\
Transformer     & 1.03  & 1.10  & 98.9  \\
Mamba           & 3.12  & 3.32   & 87.8  \\
\variant     & \textbf{1.01}  & \textbf{1.01}   & 99.8 \\
\vanilla     & \textbf{1.01}  & \textbf{1.01}   & \textbf{99.9} \\
\bottomrule
\hfill 
\end{tabular}
\caption{Perplexity (PPL) and greedy-decoding accuracy for RegBench after training each model 15 and 120 epochs. We use the same models configurations as done in \citet{Akyurek2024ICLL} and perform similar hyperparameter sweeps.  See \Cref{sec:tasks} for how accuracy is measured. * denotes reported accuracy in \citet{Akyurek2024ICLL}.}
\label{tab:icll-regbench}
\end{table}

\begin{highlight}
\paragraph{Finding 9:} 
\emph{Hybrid models perform as well as, or outperform, Transformer and Mamba in formal language ICL, as exemplified in Tables~\ref{tab:ginc} and \ref{tab:icll-regbench}.}
\end{highlight}

On GINC, Mamba achieves the best ICL accuracy among non-LSTM models, though Transformer achieves lower perplexity. 
Interestingly, \vanilla performs on par with Transformer and Mamba, while \variant performs slightly worse than other models here.
However, findings from \citet{Xie2021Explanation} indicate that LSTMs excel over Transformers on GINC, even when accounting for different settings such as vocabulary size or the number of in-context examples.
This aligns with previous findings in which Transformers perform worse or comparably to LSTMs in many formal languages considered~\citep{Bhattamishra2020FormalLanguages, Deletang2022Neural}.
Yet, Transformers are the \emph{de facto} superior model for language modeling, so it remains unclear how performance on this benchmark translates to real-world language ICL, where Transformers typically outperform LSTMs.

On RegBench, which favors Transformers over attention-free models, Mamba indeed performs worse than Transformer, consistent with previous findings. Notably, hybrid architectures excel on this benchmark, converging much faster both Mamba and Transformer while achieving higher accuracy. 

Given prior evidence that \vanilla achieves lower perplexity in language modeling~\citep{Gu2023mamba}, our new results suggest that hybrid models offer a promising direction for both language modeling and in-context learning on language tasks. We hope these results and analysis demonstrate the potential of hybrid models for language-based applications of ICL.
\section{Discussion}\label{sec:conclusion}

In this work, we have provided a comprehensive investigation of in-context learning with state-space models (SSMs) and contrasted them with the Transformer architecture. Our study has revealed that SSMs, especially Mamba, are capable in-context learners. On the other hand, our evaluations revealed that neither SSMs nor Transformers are great at all tasks: SSMs struggle with decision tree and retrieval tasks whereas Transformers struggle with sparse parity. This has led us to the hybrid architecture \variant which achieves a best-of-both-worlds performance on our ICL suite.

Future research directions include exploring (1) how the performance on our ICL suite correlates with general language modeling capabilities, such as perplexity on standard NLP benchmarks, (2) developing more effective architectures by integrating elements from transformers, SSMs, and gating mechanisms, (3) identifying architectural features that contribute to effective in-context learning, and (4) assessing the impact of \variant and other innovative architectures on language modeling performance.

\section*{Impact Statement}

This paper provides a comprehensive study of language modeling architectures which help identify their weaknesses, strengths, and provide recipes for new architectures. The outcomes of this work will potentially facilitate efficiency and architectural improvements for large language models.


\section*{Acknowledgement}
The work of Dimitris Papailiopoulos is supported in part by ONR Grant No. N00014-21-1-2806 and No. N00014-23-1-2848. The work of Samet Oymak is supported in part by NSF CAREER Award CCF-2046816. The work of Jaeseung Park was supported by KRAFTON AI Fellowship. The authors would like to thank Byeongju Kim and Seongjun Yang for helpful discussion and Gibbeum Lee for valuable feedback on an early draft of this paper.

\bibliography{main}
\bibliographystyle{icml2024}

\newpage
\appendix
\onecolumn

\section{Experimental Setup}\label{appendix:setup}

In this section, we describe our experimental design and configured setup. Our code and detailed implementations can be found in \url{https://github.com/krafton-ai/mambaformer-icl}.

\subsection{Model architectures}

We focus on decoder-only Transformer models, particularly those from the GPT-2 family \citep{radford2019language}, Mamba \citep{Gu2023mamba}, and their hybrid variants, including \vanilla and \variant configurations. These models are evaluated across a range of sizes, as detailed in Table~\ref{tab:tf_model_conf}. Transformer layers consist of a Multi-Head Attention (MHA) block followed by a Multilayer Perceptron (MLP) block. Mamba models consist of two Mamba blocks per layer. The hybrid variants merge these approaches, combining a single MHA block with a Mamba block. For MHA blocks, we use 8 number of heads. Refer to Figure~\ref{fig:model_arch} for a visualization of the architectures considered.

\subsection{Model training and configurations}\label{appendix:model_training}

\begin{table}[h]
\centering
\begin{tabular}{lcccc}
\toprule
   Size group & (\# layers, embed dim) & Transformer & Mamba & S4* \\ \hline
Large & \{(12, 768)\} & 86M & 90M & 88M \\ 
Medium & \{(8, 512), (32, 256)\}  & 25M & 27M & 26M \\
Small & \{(4, 256), (16, 128)\} & 3M & 3M & 3M \\
X-Small  & \{(2, 128), (8, 64), (32, 32)\} & 420K & 460K & 430K\\ \toprule
&  & S4-Mamba & \vanilla & \variant \\ \hline
Large & \{(12, 768)\} & 86M  & 74M & 77M\\ 
Medium & \{(8, 512), (32, 256)\}  & 26M & 22M & 24M   \\
Small & \{(4, 256), (16, 128)\} & 3M & 3M & 3.2M \\
X-Small  & \{(2, 128), (8, 64), (32, 32)\} & 430K & 400K & 480K
\\ \bottomrule
\hfill
\end{tabular}
\caption{
    The four size groups of model architectures we have used for our experiments. For each size group, we run various learning rates in addition to training `narrower and deeper' models of the same size and FLOPs. We keep the number of heads fixed at 8. We do not train models deeper than 32 layers, as we have observed that accuracy is best in between 4 to 32 layers according to~\Cref{appendix:scale_linear}. *For S4 models, the embedding dimensions were multiplied by a factor of 1.75 to match parameters.
}\label{tab:tf_model_conf}
\end{table}

We train all of our models on A100-SXM4-40GB GPUs for 500,000 training steps on all tasks, except for vector-valued MQAR, in which the models were trained for 300,000 training steps. We use Adam optimizer~\citep{kingma2014adam} with a fixed learning rate. The default learning rate is set to $1e-4$. We also search various learning rates in $\{5e-5, 2e-4, 4e-4\}$. We observe that the training procedure is the most sensitive to choosing the right learning rate. In particular, as the number of parameters of the models increases, the training procedure is prone to gradient explosions, especially in Mamba and hybrid architecutres. Hence, we clip the gradient norm, with values in $\{5.0, 10.0, 50.0\}$.

As for the train and test data, we fix the dimension of $\bvec x$ to be $20$, and fix the batch size to be $64$. As suggested in \citet{Garg2022Transformers}, we also observe that curriculum is helpful in certain ICL tasks. We adopt a curriculum; every 2000 steps, we increase both  the dimension of $\bvec x$ and the number of points within the input context.

For model configurations, we mainly follow the four size groups of Transformers listed below (\cref{tab:tf_model_conf}). As explained in \cref{fig:model_arch}, we denote 2 blocks of Mamba, multi-head attention, or a feed forward network as 1 layer. This roughly aligns the number of parameters in Transformer, Mamba and S4-Mamba. For \vanilla and \variant, we follow the same design. This yields in models with less parameters from the lack of feed forward networks. However, both models showed strong performance to other models and hence the model configurations were kept as is. For S4, however, we further increase the embedding dimension by a factor of $1.75$ to match the number of parameters of transformers.

\new{As demonstrated, trading embedding dimension (narrow width) for additional layers (greater depth) allows for diverse model configurations while maintaining the same total parameter count. We conducted an ablation study on linear regression to explore the impact of width versus depth on ICL tasks and to identify the optimal configurations for peak performance, given the same FLOPs. For the rationale behind our experimental choices in these configurations, refer to \Cref{appendix:scale_linear} as detailed in \Cref{tab:tf_model_conf} for ICL tasks. Furthermore, in \Cref{tab:tf_model_conf}, we also explore whether the choice of optimizers affect our results and alter our conclusions.}

\subsection{Chain-of-Thought-I/O settings}

Table~\ref{tab:cot_config} presents the configurations for the Chain-of-Thought-I/O task using a 2-layer ReLU neural network, following the setup described by~\citet{Li2023DissectingCoT}. In the model scale experiment, the input dimension $d=10$ and hidden layer dimension $k=8$ are held constant while varying the model scale. Additionally, the hidden dimension $k$ is varied among ${4,8,16}$ while fixing the model scale to small to identify the effect of problem scale. 

\begin{table}[ht]
\centering
\begin{tabular}{lccc}
\toprule
Model     & \# layers & embed dim & \# heads (MHA) \\ \hline
standard & 12        & 256           & 8     \\ 
small & 6         & 128           & 4    \\
tiny & 3         & 64           & 2     \\
\bottomrule
\hfill
\end{tabular}
\caption{
    Model configurations for Chain-of-Thought-I/O experiments in \Cref{fig:CoT-IO_result}.
}
\label{tab:cot_config}
\end{table}

\subsection{Many-outlier regression settings}

We run many-outlier regression on two size groups listed below in \Cref{tab:many_config}. The configurations below required multi-GPU training due to its long context length of 1024 ($N=512$).

\begin{table}[ht]
\centering
\begin{tabular}{lccc}
\toprule
Model size     & \# of layers & Embed dim & \# of heads (MHA) \\ \hline
Regular & 6        & 512           & 8     \\ 
Mini & 4         & 256           & 8    \\
\bottomrule \hfill
\end{tabular}
\caption{
    Model configurations for many-outlier regression ICL in \Cref{fig:ortho_result}.
}
\label{tab:many_config}
\end{table}

\subsection{Vector-valued MQAR}
\label{appendix:mqar_setup}

The training set consists of 300,000 training samples. We train for $64$ epochs with batch size of $64$ and evaluate on a test set of 3,000 samples. For each setting, we sweep with learning rates in np.logspace(-4, -2, 4) and report the best result among all learning rates.

\section{FLOPs Computation}
\label{appendix:flops}

\begin{table}[ht]
\centering

\begin{tabular}{cc}
\toprule
     & Number of multiplications  \\ \midrule
QKV projection & $3LD^2$\\ 
Outer product and multiply $V$ & $2L^2D$\\
Outer projection & $LD^2$\\ \midrule
FFN with \texttt{ffw\_width=4} & $8LD^2$\\
\bottomrule \hfill
\end{tabular}
\caption{Number of multiplications in a Transformer block. $L$ denotes the input sequence length and $D$ denotes the hidden dimension of the model.}
\label{tab:tf_flops}

\vspace{5mm} 

\begin{tabular}{cc}
\toprule
     & Number of multiplications  \\ \midrule
Input projection & $2LED^2$\\ 
SSM  & $7LEDN+4LED$\\ 
Output projection & $LED^2$\\
\bottomrule \hfill
\end{tabular}
\caption{Number of multiplications in a Mamba block. $L,D$ are the same as \Cref{tab:tf_flops}. $N$ represents the state size of the SSM and $E$ denotes the expansion factor of the hidden dimension within each Mamba block. We assume $E=2$ and $N=16$.}
\label{tab:mamba_flops}

\end{table}

We count the number of multiplications in a Mamba block and a Transformer block in \cref{tab:tf_flops} and \cref{tab:mamba_flops}. We assume batch size $B=1$. To calculate FLOPs, we follow the similar methodology used in \citet{Kaplan2020Scaling, Muennighoff2023Scaling} and multiply the number of multiplications by $6$ to account for the multiply-accumulate cost in both forward and backward pass. Note that a \vanilla block is an attention block stacked with a Mamba block, so the number of multiplications in a \vanilla block is $10LD^2+2L^2D$, ignoring the linear terms.

\section{Exploring effects of width vs. depth in ICL}
\label{appendix:scale_linear}

\begin{figure*}[ht]

  \begin{subfigure}[t]{\textwidth}
    \centering
    \includegraphics[width=0.65\textwidth]{arxiv_version/imgs/ssm/legend_ssm.pdf}
    \centering
  \end{subfigure}
  \vspace{0.2mm}

  \centering
  \begin{subfigure}[b]{0.49\textwidth}
      \includegraphics[width=\textwidth]{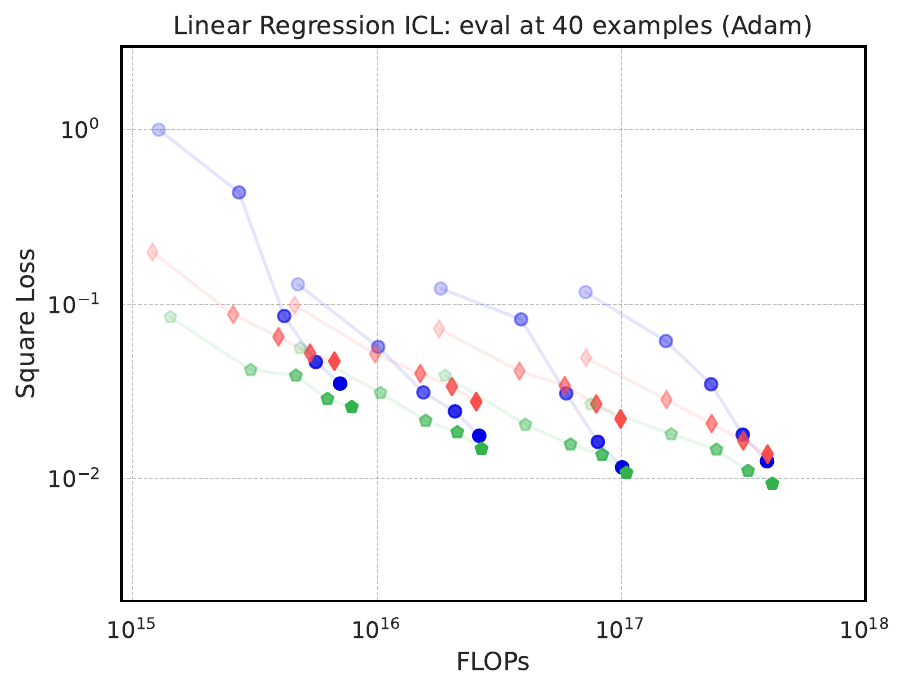}
  \end{subfigure}
  \begin{subfigure}[b]{0.49\textwidth}
      \includegraphics[width=\textwidth]{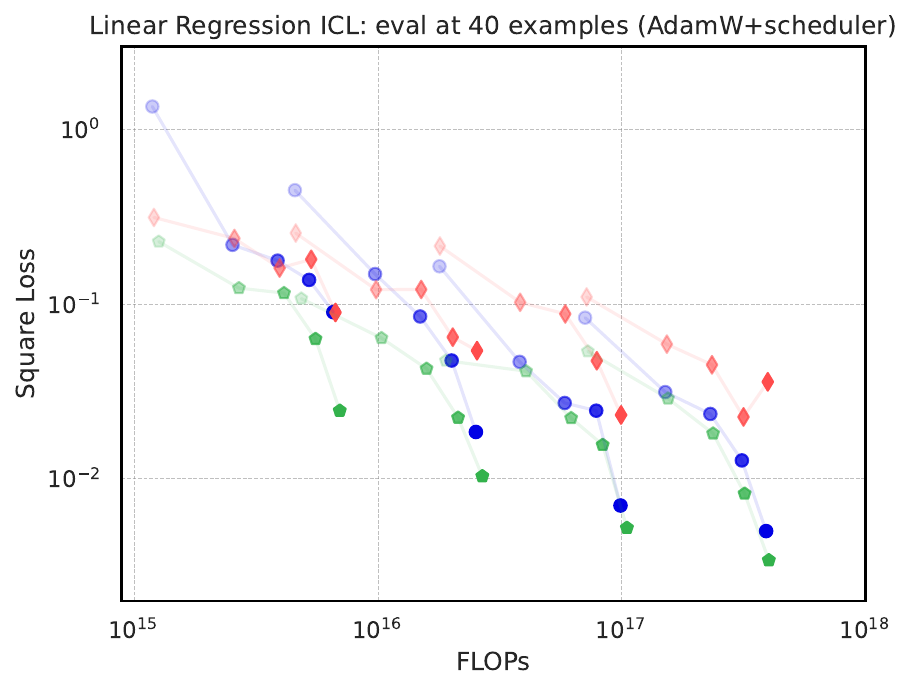}
  \end{subfigure}
  \caption{Performance of Transformer, Mamba, and \variant on linear regression ICL, using two different optimizers. The left side shows results obtained using the Adam optimizer, while the right side shows results obtained using AdamW coupled with linear warmup and cosine decay scheduler. We observe that Mamba and \variant gain considerably more than Transformer when following the new optimizer setup. See \Cref{appendix:scale_linear} for implementation details.}
  \label{fig:linreg_depth_scaling}
\end{figure*}

\new{
In this section, we empirically validate our experimental design and setup, detailed in \Cref{appendix:setup}, through a comprehensive comparison of wide and shallow networks versus deep and narrow networks, to investigate the effect of model design while fixing the total number of FLOPs. Additionally, we investigate whether comparing different architectures with Adam only is a fair comparison. We evaluate the choice of optimizers, specifically comparing Adam--used in our main experiments--with AdamW~\citep{Loshchilov2018Decoupled} coupled with a linear warmup and cosine decay scheduler, which is the standard in language model pretraining for Transformer-based models~\citep{Touvron2023Llama}.
}

\begin{figure*}[ht]
  \centering
  \begin{subfigure}[b]{0.49\textwidth}
      \includegraphics[width=\textwidth]{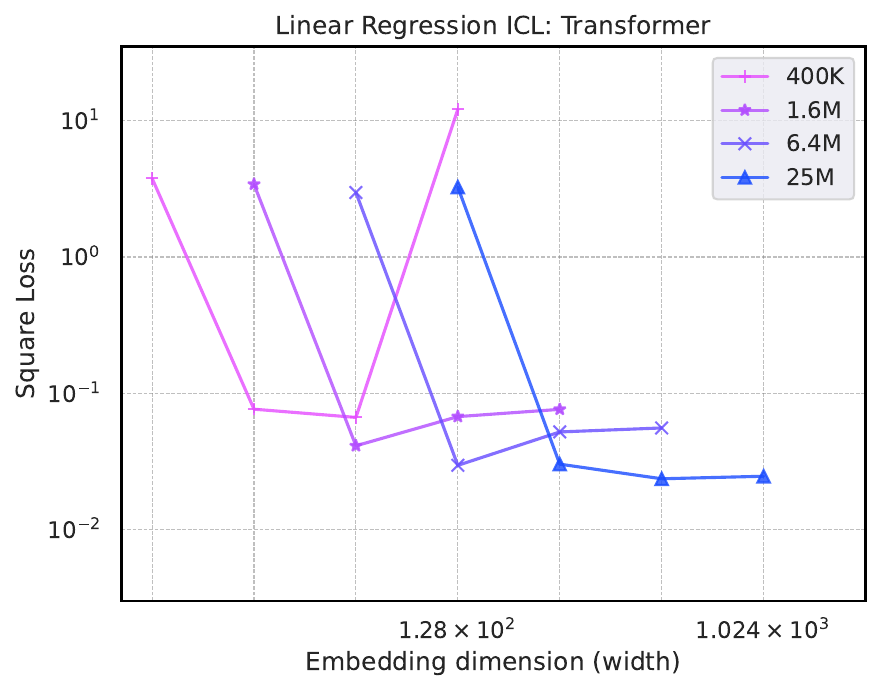}
  \end{subfigure}
  \begin{subfigure}[b]{0.49\textwidth}
      \includegraphics[width=\textwidth]{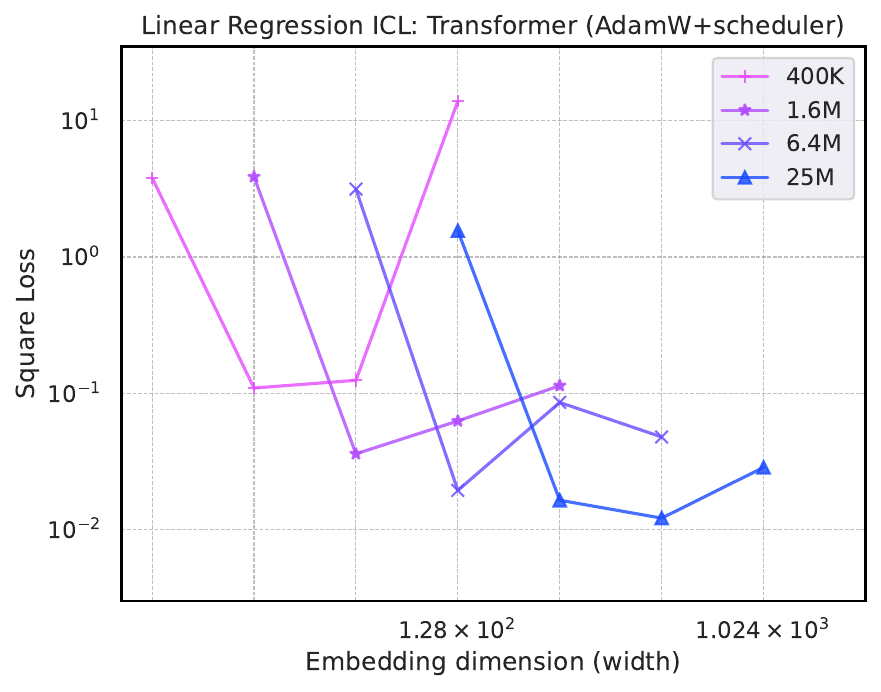}
  \end{subfigure}
  \begin{subfigure}[b]{0.49\textwidth}
      \includegraphics[width=\textwidth]{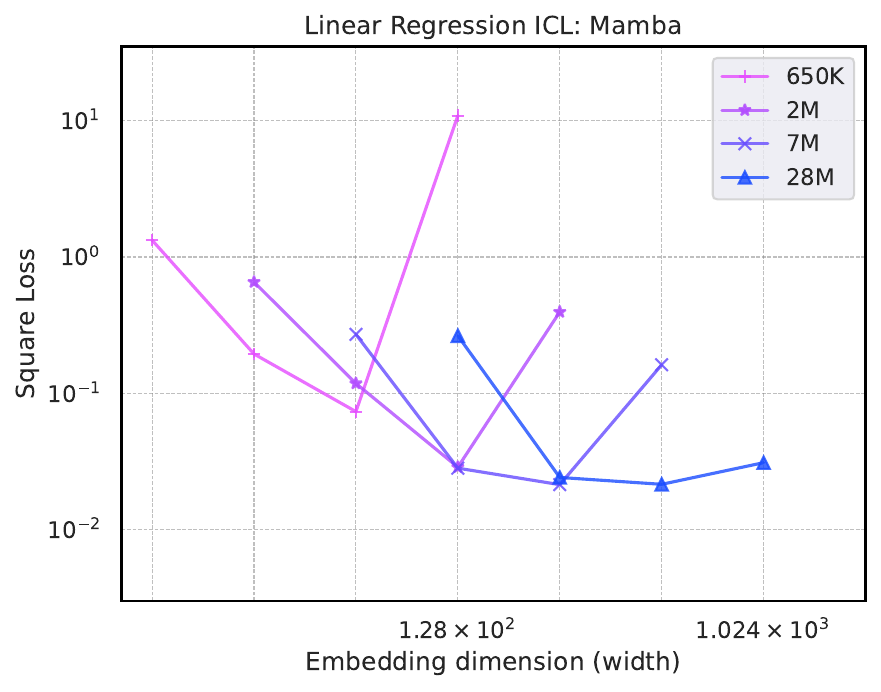}
  \end{subfigure}
  \begin{subfigure}[b]{0.49\textwidth}
      \includegraphics[width=\textwidth]{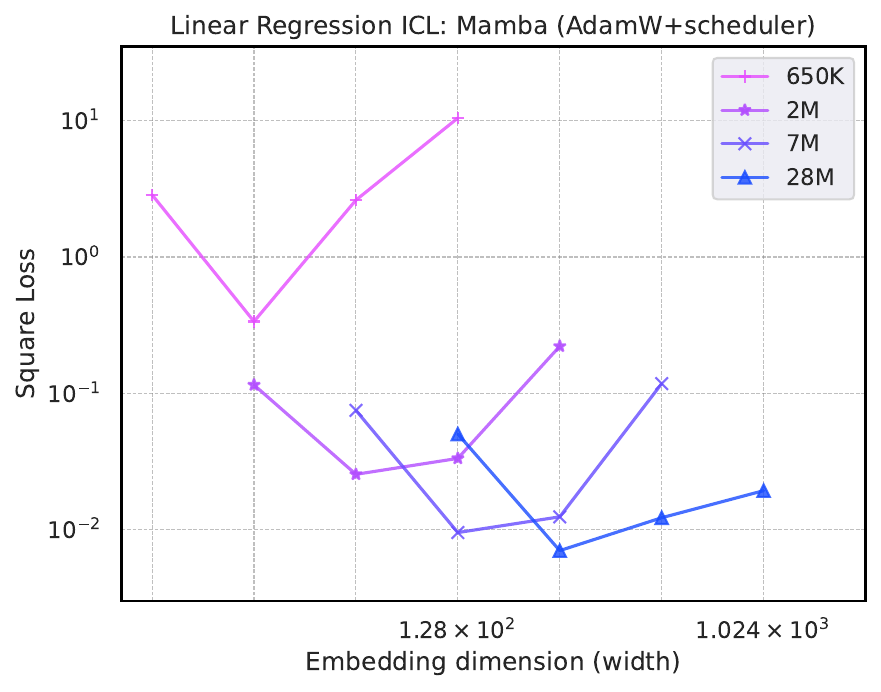}
  \end{subfigure}
  \begin{subfigure}[b]{0.49\textwidth}
      \includegraphics[width=\textwidth]{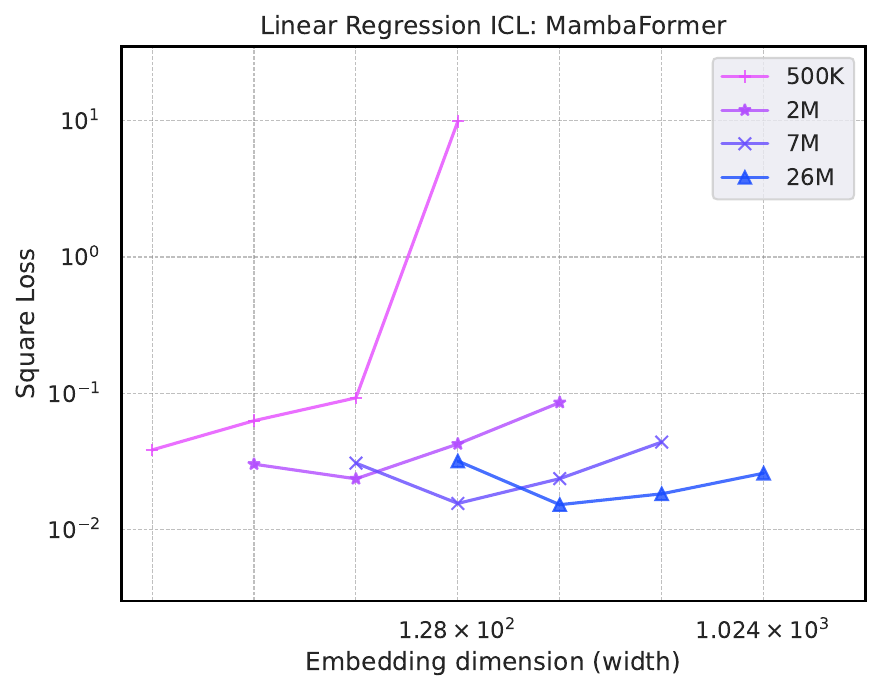}
  \end{subfigure}
  \begin{subfigure}[b]{0.49\textwidth}
      \includegraphics[width=\textwidth]{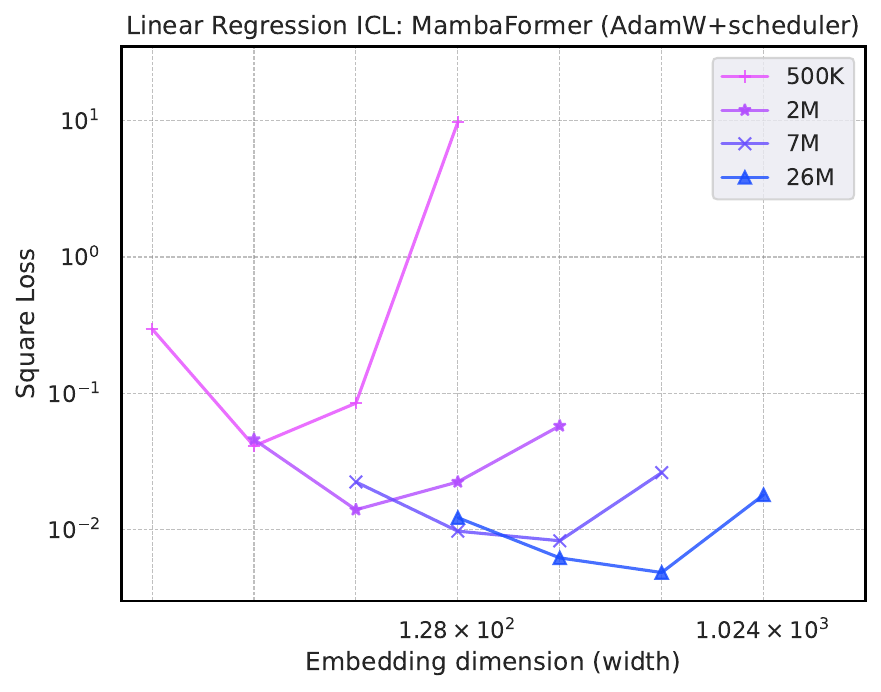}
  \end{subfigure}
  \caption{Performance of models with varying configurations on linear regression ICL eval at 40 examples. Each color of the legend represents a model group of fixed total number of (non-embedding) parameters. The models differ in the number of layers, specifically 2, 8, 32, and 128 layers for each connected line, \textit{i.e.}, the model with the widest width has 2 layers, and the model with the narrowest width has 128 layers. \textbf{(Left)} Adam optimizer as used in \Cref{sec:training}. \textbf{(Right)} AdamW with linear warmup and cosine decay.}
  \label{fig:linreg_depth}
\end{figure*}

\paragraph{Experiment settings.}
We conduct our ablation study on the task of linear regression. We explore four size groups of models, each size group with four different configurations for width and depth as seen in \Cref{fig:linreg_depth}. We study the three models of interest: Transformer, Mamba, and \variant. For Transformer and \variant, we keep the dimension of each head constant at 16; for instance, a model with embedding dimension 256 has a total of 16 heads.

For the new optimizer, we use AdamW with the following hyperparameters: $\beta_1 = 0.9, \beta_2 = 0.95$.
The final learning rate of the cosine decay scheduler is equal to 10\% of the predefined learning rate. We use linear warmup for the first 50,000 steps of training, starting from 10\% of the learning rate. For each model configuration, we choose the best performance out of learning rates $\{5e-5, 1e-4, 2e-4\}$. The remaining experiment details are identical to those described in \Cref{sec:experiments}.

\begin{highlight}
\paragraph{Finding 10:}
\emph{Shallow models struggle to learn regression, especially for Transformers. Very deep models also struggle, but less so if sufficient width is provided.}
\end{highlight}

For Transformers, we observe that 2-layer models struggle with learning linear regression, even after seeing 40 in-context examples, regardless of depth (see \Cref{fig:linreg_depth}). It appears that Transformers have a minimum layer threshold to effectively learn regression tasks. While less pronounced, a similar pattern is observed with Mamba; however, \variant demonstrates proficiency in learning regression with as few as 2 layers. Conversely, models that are excessively deep, with 128 layers, face difficulties in learning when their width is insufficient, and they also require significantly more training time due to prolonged forward and backward passes. Consequently, exploring layer counts between 4 and 32 on other tasks seems to be sufficient in identifying the optimal model configuration given fixed total FLOPs, as detailed in \Cref{tab:tf_model_conf}.

\begin{highlight}
\paragraph{Finding 11:}
\emph{AdamW coupled with a scheduler improves performance, especially for larger models and for Mamba and \variant. Yet, our conclusion remains the same and is agnostic to the choice of optimizer.}
\end{highlight}

As seen in \Cref{fig:linreg_depth_scaling} and \Cref{fig:linreg_depth}, AdamW combined with a learning rate scheduler helps learning stronger models, as one may expect from its wide adoption in large language model training. 
The benefits of the new optimizer increase with model size across different architectures. However, we note that Mamba and MambaFormer gain more from AdamW plus a scheduler compared to Transformer. Anecdotally, the learning rate schedule is particularly beneficial for training Mamba, as deeper and larger Mamba models tend to experience gradient explosion issues.

The two primary conclusions from our work were: (1) Mamba is capable of performing ICL effectively, and (2) MambaFormer successfully combines high performance with the best attributes of both Transformer and Mamba, sometimes even surpassing them. 
Our empirical tests suggest that the choice of optimizer and scheduler does not fundamentally alter these conclusions. In fact, our ablation study indicates that Mamba and MambaFormer might perform even better under these conditions. Therefore, we decided to use the Adam optimizer in our main experiments, consistent with prior work~\citep{Garg2022Transformers, Bhattamishra2023TransformersLLMs} on Transformer ICL.

\end{document}